\documentclass[10pt,twocolumn,letterpaper]{article}
\usepackage[pagenumbers]{iccv}

\usepackage{algorithm}
\usepackage{listings}
\usepackage{graphicx}
\usepackage{amsmath}
\usepackage{amssymb}
\usepackage{booktabs}
\usepackage{algpseudocode}
\usepackage[accsupp]{axessibility} 
\usepackage{etoolbox}
\makeatletter
\AfterEndEnvironment{algorithm}{\let\@algcomment\relax}
\AtEndEnvironment{algorithm}{\kern2pt\hrule\relax\vskip3pt\@algcomment}
\let\@algcomment\relax
\newcommand\algcomment[1]{\def\@algcomment{\footnotesize#1}}
\renewcommand\fs@ruled{\def\@fs@cfont{\bfseries}\let\@fs@capt\floatc@ruled
  \def\@fs@pre{\hrule height.8pt depth0pt \kern2pt}%
  \def\@fs@post{}%
  \def\@fs@mid{\kern2pt\hrule\kern2pt}%
  \let\@fs@iftopcapt\iftrue}
\makeatother

\usepackage{times}
\usepackage{epsfig}
\usepackage{graphicx}
\usepackage{amsmath}
\usepackage{amssymb}
\usepackage{verbatim}
\usepackage{hhline}
\usepackage{multirow}
\usepackage{makecell}
\usepackage{booktabs}
\usepackage{algpseudocode}

\usepackage{algorithm}
\usepackage{listings}
\usepackage{xcolor}
\usepackage{caption}
\usepackage{tikzducks}

\usepackage{ amssymb }
\makeatletter
\newcommand{\ostar}{\mathbin{\mathpalette\make@circled\star}}
\newcommand{\make@circled}[2]{%
  \ooalign{$\m@th#1\smallbigcirc{#1}$\cr\hidewidth$\m@th#1#2$\hidewidth\cr}%
}
\newcommand{\smallbigcirc}[1]{%
  \vcenter{\hbox{\scalebox{0.77778}{$\m@th#1\bigcirc$}}}%
}
\makeatother

\makeatletter
\newcommand\footnoteref[1]{\protected@xdef\@thefnmark{\ref{#1}}\@footnotemark}
\makeatother

\usepackage{xcolor,colortbl}

\definecolor{grey}{rgb}{0.9, 0.9, 0.9}
\definecolor{electricviolet}{rgb}{0.56, 0.0, 1.0}

\usepackage{xspace}

\usepackage{amssymb}
\usepackage{pifont}

\makeatletter
\DeclareRobustCommand\onedot{\futurelet\@let@token\@onedot}
\def\@onedot{\ifx\@let@token.\else.\null\fi\xspace}
\def\eg{\emph{e.g}\onedot} 
\def\ie{\emph{i.e}\onedot}

\newcommand{\rcol}{\rowcolor{grey}}

\newcommand{\ours}{DiffGEBD}

\newcolumntype{L}[1]{>{\raggedright\let\newline\\\arraybackslash\hspace{0pt}}m{#1}}
\newcolumntype{C}[1]{>{\centering\let\newline\\\arraybackslash\hspace{0pt}}m{#1}}
\newcolumntype{R}[1]{>{\raggedleft\let\newline\\\arraybackslash\hspace{0pt}}m{#1}}

\usepackage{amsmath,amsfonts,bm}

\def\eqref#1{equation~\ref{#1}}

\def\1{\bm{1}}

\def\vy{{\bm{y}}}

\DeclareMathAlphabet{\mathsfit}{\encodingdefault}{\sfdefault}{m}{sl}
\SetMathAlphabet{\mathsfit}{bold}{\encodingdefault}{\sfdefault}{bx}{n}

\newcommand{\E}{\mathbb{E}}

\definecolor{iccvblue}{rgb}{0.21,0.49,0.74}
\usepackage[pagebackref,breaklinks,colorlinks,allcolors=iccvblue]{hyperref}

\def\paperID{1259}
\def\confName{ICCV}
\def\confYear{2025}

\title{Generic Event Boundary Detection via Denoising Diffusion}

\author{
Jaejun Hwang\textsuperscript{1,2}\thanks{Equal contribution.} \quad 
Dayoung Gong\textsuperscript{1}\footnotemark[1] \quad 
Manjin Kim\textsuperscript{1} \quad 
Minsu Cho\textsuperscript{1} \quad \\ \\
\textsuperscript{1}Pohang University of Science and Technology (POSTECH) \quad\quad
\textsuperscript{2}GenGenAI
 \\
\href{https://cvlab.postech.ac.kr/research/DiffGEBD}{\texttt{\small https://cvlab.postech.ac.kr/research/DiffGEBD}}
}

\begin{document}
\maketitle
\begin{abstract}
Generic event boundary detection (GEBD) aims to identify natural boundaries in a video, segmenting it into distinct and meaningful chunks. Despite the inherent subjectivity of event boundaries, previous methods have focused on deterministic predictions, overlooking the diversity of plausible solutions.
In this paper, we introduce a novel diffusion-based boundary detection model, dubbed DiffGEBD, that tackles the problem of GEBD from a generative perspective.
The proposed model encodes relevant changes across adjacent frames via temporal self-similarity and then iteratively decodes random noise into plausible event boundaries being conditioned on the encoded features. 
Classifier-free guidance allows the degree of diversity to be controlled in denoising diffusion.
In addition, we introduce a new evaluation metric to assess the quality of predictions considering both diversity and fidelity.
Experiments show that our method achieves strong performance on two standard benchmarks, Kinetics-GEBD and TAPOS, generating diverse and plausible event boundaries. 
\end{abstract}    
\vspace{-1mm}
\section{Introduction}
\label{sec:intro}

Through the intricate workings of visual perception, humans can effortlessly detect and interpret a wide range of changes in subjects, objects, and scenes.
Research in cognitive science demonstrates that the human visual system easily divides a temporal sequence of images into units of semantic significance~\cite{10.1037/0033-2909.127.1.3}.
The task of generic event boundary detection (GEBD) has recently been proposed to identify these natural event boundaries in a similar spirit~\cite{shou2021generic, lea2016segmental, Lin_2019_ICCV, kang2022uboco, tan2023temporal, li2022end, tang2022progressive, li2022structured}.
While conventional video tasks in computer vision, such as action recognition~\cite{carreira2017quo, tran2015learning, arnab2021vivit, wang2018non, tran2018closer, wu2019long}, temporal action detection~\cite{xu2020g, zeng2019graph, zhu2021enriching, kahatapitiya2021coarse, zhang2022actionformer}, and temporal action segmentation~\cite{farha2019ms, yi2021asformer, liu2023diffusion, ding2023temporal} mainly focus on identifying class labels or boundaries of predefined action classes, GEBD aims to localize more generic and class-agnostic event boundaries from a video. 

\begin{figure}[t]
     \centering
     \resizebox{0.9\linewidth}{!}{
     \includegraphics[width=\linewidth]{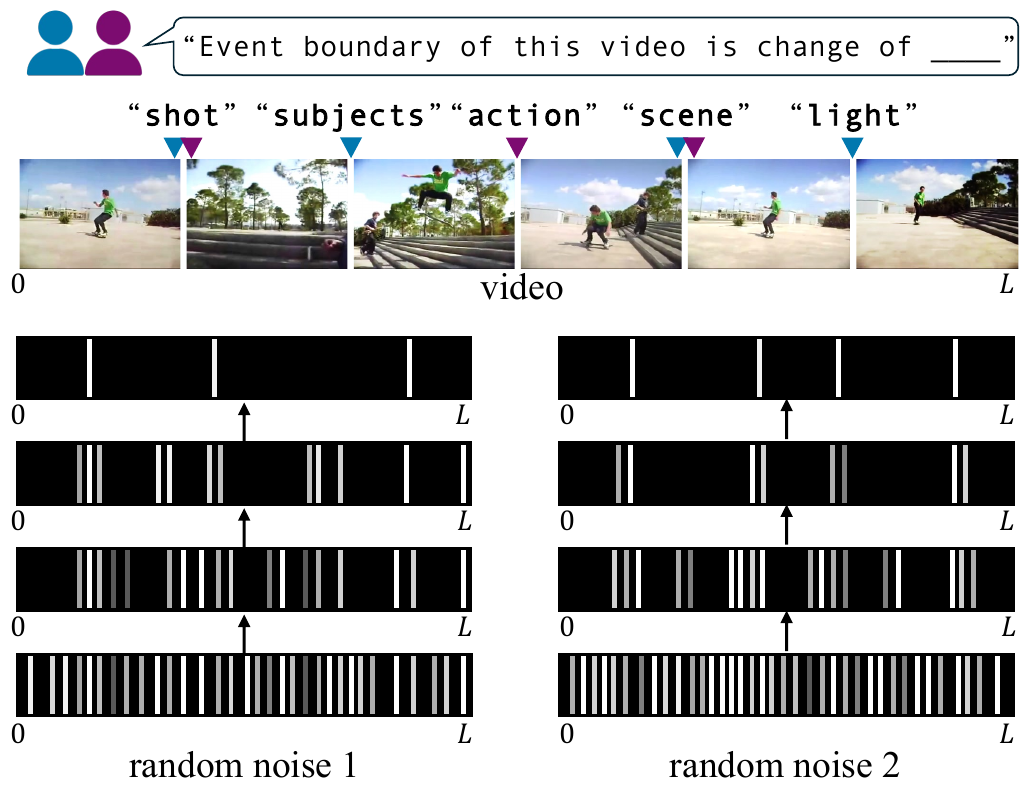}
     }
     \caption{\textbf{Generic event boundary detection from a generative perspective.} Our method generates diverse and plausible boundary predictions for generic events via denoising diffusion.} 
     \label{fig:teaser}
     \vspace{-3mm}
\end{figure}

Since generic event boundaries are inherently subjective and variable, the problem of GEBD needs to consider the diversity of human judgment; perception of these boundaries can differ significantly among individuals, leading to variation in how people identify boundaries.
To account for such human subjectivity, Kinetics-GEBD~\cite{shou2021generic}, the benchmark dataset for GEBD, provides multiple annotations from human annotators for each video. 
However, previous methods~\cite{shou2021generic, lea2016segmental, Lin_2019_ICCV, kang2022uboco, tan2023temporal, li2022end, tang2022progressive, li2022structured} all focus on predicting accurate boundaries with a deterministic model for a given video, ignoring the diversity of potential solutions. Diffusion models are well-suited for this challenge, as they naturally enable sampling diverse outputs conditioned on the same input by varying the initial noise during the stochastic denoising.

In this paper, we introduce a diffusion-based boundary detection model, dubbed \ours, which formulates the problem of GEBD from a generative perspective (Fig.~\ref{fig:teaser}).
The proposed method consists of a temporal self-similarity encoder and a denoising decoder, where the encoder captures dynamic visual changes across adjacent frames using temporal self-similarity~\cite{zheng2024rethinking}, and then the decoder iteratively denoises random noise into plausible event boundaries being conditioned on the encoded features.
Since our model outputs distinct predictions from different noises, we control the prediction diversity incorporating classifier-free guidance (CFG)~\cite{ho2021classifier} into our denoising process.
By controlling the guidance weight, DiffGEBD effectively perform diverse yet accurate boundary detection while better reflecting human judgment variability. Given the ability of our model to generate diverse predictions, a key challenge emerges in how to properly evaluate them. 

The conventional evaluation metric for GEBD, the F1 score, measures the alignment between a single prediction and multiple ground-truth annotations. However, it does not account for many-to-many alignments when a model generates multiple predictions, nor does it capture the diversity across those predictions.
To address these limitations, we introduce a diversity-aware evaluation protocol with two metrics: symmetric F1 and diversity score. The symmetric F1 captures many-to-many alignments between sets of predictions and ground-truth annotations, while the diversity score directly quantifies variation among the predictions themselves. Together, these metrics enable a more comprehensive evaluation of GEBD, better reflecting the inherent ambiguity and variability of event boundaries.

Our contributions can be summarized as follows:
\textbf{1)} we introduce a novel diffusion-based event boundary detection model, dubbed \ours, formulating GEBD from a generative perspective.
\textbf{2)} 
The degree of diversity in generated predictions can be controlled by adopting classifier-free guidance in the denoising process.
\textbf{3)} We propose a diversity-aware evaluation protocol introducing two metrics: symmetric F1 and diversity scores.
\textbf{4)} 
\ours~achieves strong performance on standard benchmark datasets, Kinetics-GEBD and TAPOS, generating diverse and plausible event boundaries.
\section{Related Work}
\label{sec:related_works}
\noindent
\textbf{Generic event boundary detection.}
GEBD~\cite{shou2021generic} is a video boundary detection task that segments a video into discrete event units, similar to how humans naturally perceive and distinguish events. 
Each event boundary marks a transition, dividing the video into shorter, taxonomy-agnostic segments. Existing approaches have primarily focused on how to effectively utilize visual information for boundary detection. UBoCo~\cite{kang2022uboco} proposes the temporal self-similarity matrix (TSM) to capture semantic inconsistency existing at video boundaries. 
DDM-Net~\cite{tang2022progressive} introduces the progressive attention to fuse spatial features and temporal similarities. LCVS~\cite{zhang2024local} further enhances boundary detection by incorporating motion vectors with similarity features. Recent approaches~\cite{li2022structured, zheng2024rethinking, zheng2025fine} improve the effectiveness of TSM by applying it in a sliding window manner over local temporal regions. All of these previous methods are deterministic, yielding a single prediction for each video. In contrast, we introduce a generative perspective to the task, which enables diverse and plausible boundary detections.

\noindent
\textbf{Diversity-aware prediction.}
Generating diverse predictions is a key challenge in tasks with inherent ambiguity, such as future action anticipation~\cite{abu2019uncertainty, zatsarynnagated} and medical image segmentation~\cite{kohl2018probabilistic,rahman2023ambiguous}. In these domains, prior works have successfully used generative models to capture the underlying data distribution. UAAA~\cite{abu2019uncertainty} proposes a framework that models probability distributions and generates multiple samples corresponding to different possible sequences of future activities. GTDA~\cite{zatsarynnagated} leverages diffusion models to capture the distribution of activities and propose a metric for measuring the diversity of generated samples. In medical image segmentation, multiple expert annotations often lead to ambiguity. Probabilistic U-Net~\cite{kohl2018probabilistic} addresses this challenge by learning to capture the distribution of annotations. They propose using Generalized Energy Distance (GED)~\cite{bellemare2017cramer, salimans2018improving, SZEKELY20131249} to evaluate the similarity between the distribution of predicted samples and annotations. In this paper, we propose a diffusion-based model that generates diverse boundary predictions for a single video and enables effective control over the degree of diversity.
Furthermore, we introduce new evaluation metrics tailored for diversity-aware prediction in the context of GEBD.

\begin{figure*}[t!]
     \centering
     \includegraphics[width=\linewidth]{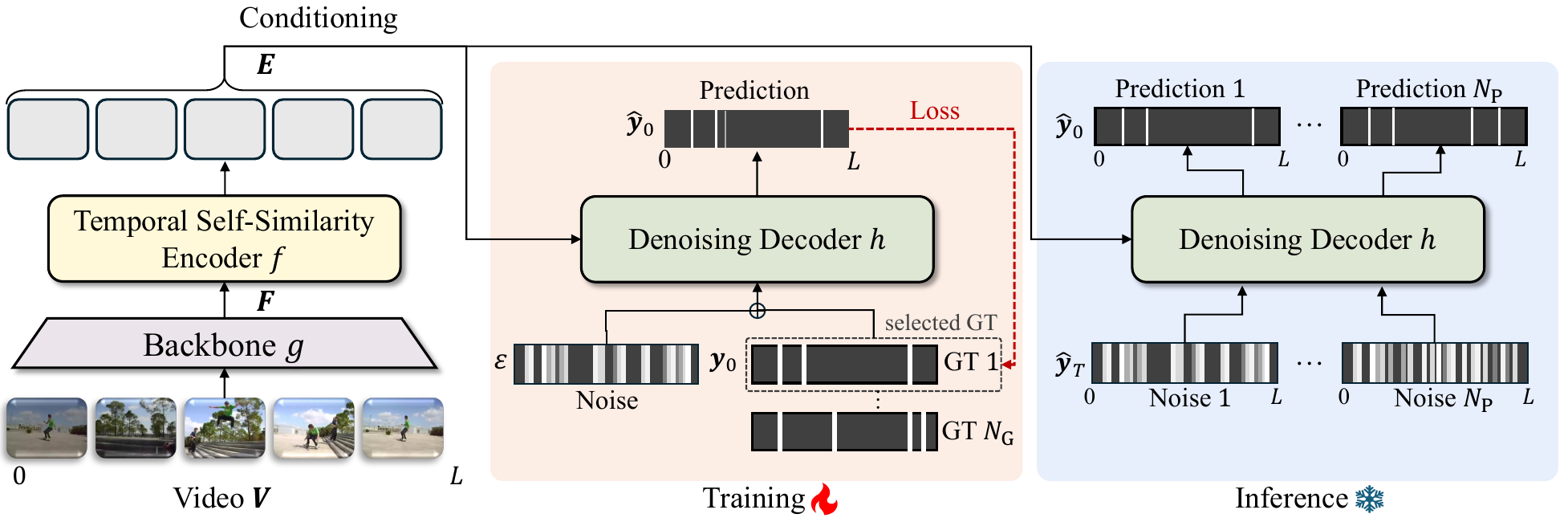}
      \vspace{-7mm}
     \caption{ \textbf{Overview of \ours.}  
     Input video $\bm{V}$ is given to the backbone network $g$, producing visual features $\bm{F}$ as output.
     Then, the extracted visual features $\bm{F}$ are produced to the encoder $f$, generating $\bm{E}$.
     During training, Gaussian noise $\epsilon$ is added to the ground-truth label $\vy_0$ following the diffusion forward step. The decoder $h$ then predicts boundaries from a noisy label $\vy_t$ at time step $t$ conditioned on $\bm{E}$. During inference, the decoder iteratively denoises starting from the random Gaussian noise $\hat{\bm{y}}_T$, generates the predictions, \textit{i.e.}, $\hat{\bm{y}}_T \rightarrow \hat{\bm{y}}_{T-\Delta} \rightarrow
 \cdots \rightarrow \hat{\bm{y}}_0$, following DDIM inference step~\cite{song2020denoising}. By differently initializing the $N_\mathrm{P}$ random Gaussian noises, we can generate $N_\mathrm{P}$ diverse predictions using a single model.
     }
     \label{fig:pipeline}
     \vspace{-4mm}
\end{figure*}
\noindent
\textbf{Diffusion model.}
The diffusion model~\cite{sohl2015deep} is a generative model inspired by non-equilibrium statistical physics~\cite{jarzynski1997equilibrium}, that learns a data distribution by reversing a gradual noising process.
The remarkable success of the diffusion model in image generation~\cite{ho2020denoising, song2020denoising, rombach2022high, ramesh2022hierarchical, nichol2022glide} and their conditional variants~\cite{ho2021classifier, dhariwal2021diffusion}
has recently extended to a range of computer vision domains, including image segmentation~\cite{amit2021segdiff, wu2024medsegdiff}, object detection~\cite{chen2023diffusiondet}, and video understanding~\cite{liu2023diffusion, foo2024action}. Following this paradigm, we introduce a diffusion-based boundary detection model that addresses the problem of GEBD from a generative perspective. 
\vspace{-2mm}
\section{Preliminary}
\label{sec:preliminary}

In this section, we provide the preliminaries on diffusion models
~\cite{ho2020denoising, sohl2015deep}. A diffusion model consists of two key components: a forward process that progressively adds Gaussian noise to the data, and a reverse process that reconstructs the original sample by iteratively denoising it.

\noindent
\textbf{Forward Process.}
The forward process involves adding small Gaussian noise $\epsilon$ over \textit{T} timesteps to create noisy data from the given real data distribution $\mathbf{x}_0 \sim q(\mathbf{x})$. The size of the noise added during the forward process is controlled by a variance schedule $\{\beta_t \in (0, 1)\}_{t=1}^T$. Let $\alpha_t = 1-\beta_t$ and $\bar{\alpha}_t = \prod_{i=1}^t \alpha_i$. The forward process is defined as:
\begin{align}
  q(\mathbf{x}_t | \mathbf{x}_{t-1}) &= \mathcal{N}(\mathbf{x}_t; \sqrt{\alpha_t} \mathbf{x}_{t-1}, \alpha_t \mathbf{I}),\\
q(\mathbf{x}_{1:T} | \mathbf{x}_0) &= \prod_{t=1}^T q(\mathbf{x}_t | \mathbf{x}_{t-1}).  
\end{align}
By the Markovian chain, this can be formulated as follows:
\begin{equation}
q(\mathbf{x}_t|\mathbf{x}_0) = \mathcal{N}(\mathbf{x}_t|\sqrt{\bar{\alpha_0}}\mathbf{x}_0, (1-\bar{\alpha_0}\textbf{\textit{I}})).
\end{equation}
Finally, with the reparameterization trick, we obtain:
\begin{equation}
\mathbf{x}_t = \sqrt{\bar{\alpha}_t} \mathbf{x}_0 + \epsilon \sqrt{1 - \bar{\alpha}_t}, \quad \epsilon \sim \mathcal{N}(0,  \textbf{\textit{I}}).
\label{eq:diffusion_forward}
\end{equation}

\noindent
\textbf{Reverse Process.}
The reverse process estimates \( \mathbf{x}_0 \) from \( \mathbf{x}_t \), inverting the forward process.
This requires estimating the posterior  \( q(\mathbf{x}_{t-1} | \mathbf{x}_t) \), which is intractable as it depends on the real data distribution $q(x_0)$.
Therefore, we approximate the posterior with a learned model distribution \( p_\theta(\mathbf{x}_{t-1} | \mathbf{x}_t) \): 
 \begin{equation}
p_{\theta}(\mathbf{x}_{t-1} | \mathbf{x}_t) = \mathcal{N}(\mathbf{x}_{t-1}; \mu_{\theta}(\mathbf{x}_t, t), \sigma^2_tI),   
\end{equation}
where $\mu_{\theta}(\mathbf{x}_t, t)$ is a predicted mean parameterized by deep neural network, and $\sigma^2_t$ is a variance term determined by $\beta_t$.
Instead of predicting \(\mu_{\theta}(\mathbf{x}_t, t)\) directly, we let the model predict \(\mathbf{x}_0\) by neural network $f_\theta (\mathbf{x}_t, t)$. Starting from pure random noise \( \mathbf{x}_T \), the model can reduce the noise using the following update rule:
\begin{align}
\label{eq:diffusion_backward}
\mathbf{x}_{t-1} = &\sqrt{\bar{\alpha}_{t-1}} f_{\theta}(\mathbf{x}_t, t) + \nonumber\\
& \sqrt{1 - \bar{\alpha}_{t-1} - \sigma_t^2} \cdot \frac{\mathbf{x}_t - \sqrt{\bar{\alpha}_t} f_{\theta}(\mathbf{x}_t, t)}{\sqrt{1 - \bar{\alpha}_t}} + \sigma_t \epsilon
\end{align}
By iteratively applying Eq.~\ref{eq:diffusion_backward}, the model can generate samples from $p_{\theta}$ via a trajectory from $T$ to $0$. DDIM sampling~\cite{song2020denoising} improves efficiency by skipping intermediate steps,
\textit{i.e.}, $\textbf{x}_T \rightarrow \textbf{x}_{T-\Delta} \rightarrow \ldots \rightarrow \textbf{x}_0$.

\section{Proposed Approach}
\label{sec:method}
We introduce DiffGEBD, a novel diffusion-based framework for generic event boundary detection. This section provides the problem setup (Sec.~\ref{sec:problem_setup}), details of \ours~(Sec.~\ref{sec:DiffGEBD}), the training objective (Sec.~\ref{sec:training_object}), and integration of the classifier-free guidance(Sec.~\ref{sec:cfg}).

\subsection{Problem setup}
\label{sec:problem_setup}
Given a video $\bm{V} \in \mathbb{R}^{L \times H \times W \times 3}$ consisting of $L$ frames, where each frame has height $H$, width $W$, and RGB channels, the goal of generic event boundary detection (GEBD) is to identify a sequence of event boundaries $\bm{y} \in \{0,1\}^{L}$. Each element $\bm{y}_l$ is a binary indicator that represents whether an event boundary is present, with 1 indicating presence and 0 indicating absence at frame $l$.

\subsection{DiffGEBD}
\label{sec:DiffGEBD}
The overall architecture of \ours~is illustrated in Fig.~\ref{fig:pipeline}.
The input video is fed into a backbone network $g$ to extract visual feature representations. The encoder $f$ captures relevant temporal changes across adjacent frames via temporal self-similarity, and the denoising decoder $h$ refines random Gaussian noise into event boundary predictions conditioned on the visual embeddings produced by the encoder.  

During training, we randomly sample a diffusion time step $t\in\{1, 2, \dots, T\}$ and add noise $\epsilon \sim \mathcal{N}(0, \bm{I})$ to the ground-truth boundary label $\bm{y}_0$  following Eq.~\ref{eq:diffusion_forward}, generating noisy boundary label $\bm{y}_t$ at time step $t$.
The decoder takes $\bm{y}_t$ as input and is trained to reconstruct the original boundary label $\bm{y}_0$.
For each video with $N_\mathrm{G}$ ground-truth (GT) annotations, we select one annotation per iteration to serve as the GT, ensuring that every annotation is used once per epoch.

During inference, the decoder starts from Gaussian noise $\hat{\bm{y}}_T$ and progressively denoises it through multiple steps, \textit{i.e.}, $\hat{\bm{y}}_T \rightarrow \hat{\bm{y}}_{T-\Delta} \rightarrow
 \cdots \rightarrow \hat{\bm{y}}_0$, following DDIM sampling procedure~\cite{song2020denoising}.
Here, $\hat{\bm{y}}$ denotes predicted boundaries.
For diverse and plausible predictions, \ours~can generate $N_\mathrm{P}$ predictions with a single model by randomly initializing the starting Gaussian noise $\hat{\bm{y}}_T$ for each prediction.

\begin{figure}[t]
     \centering
     \resizebox{0.9\linewidth}{!}{
     \includegraphics[width=\linewidth]{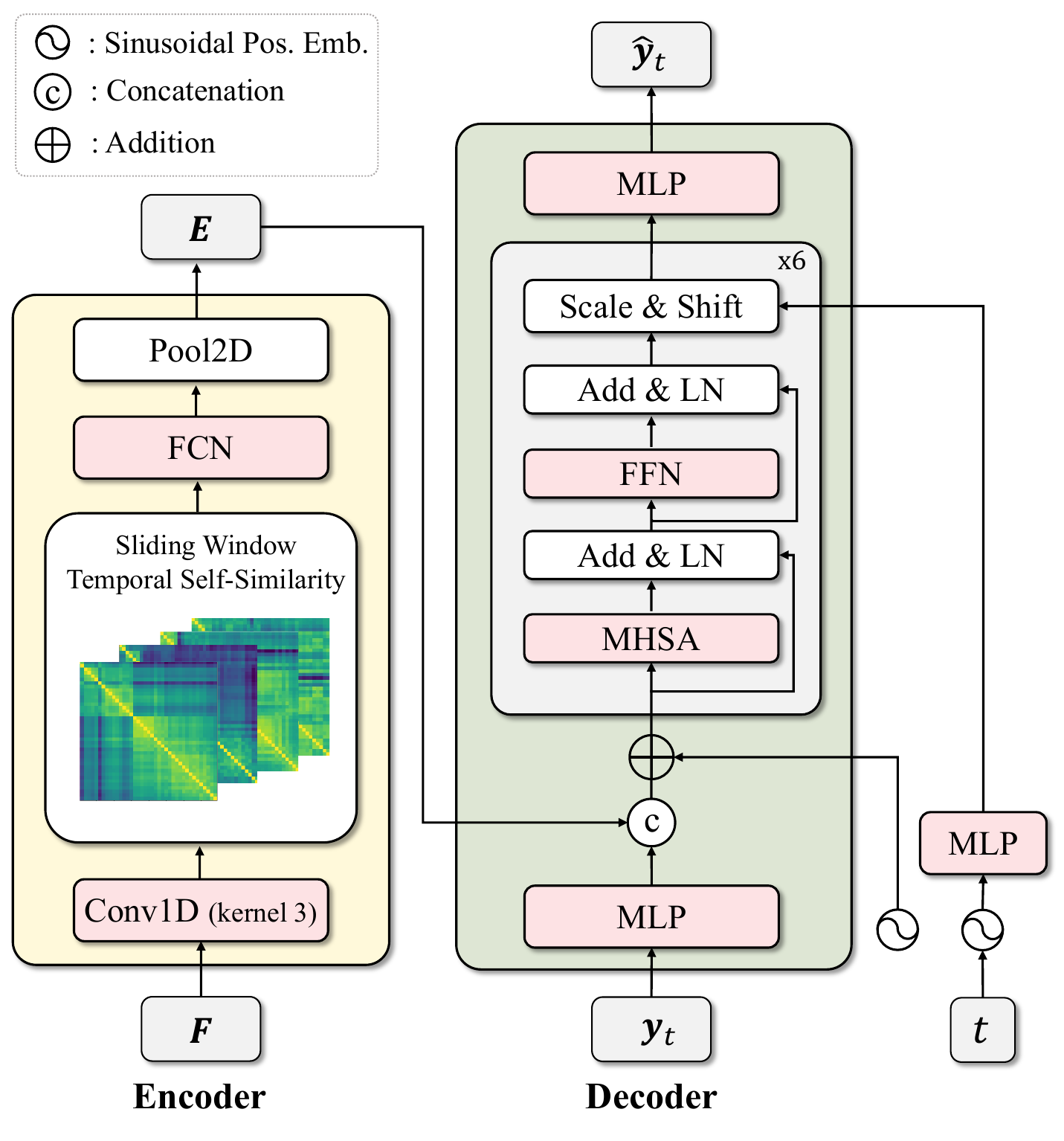}
     }
     \caption{\textbf{Detailed architecture of encoder and decoder}}
     \label{fig:architecture}
     \vspace{-5mm}
\end{figure}

\paragraph{Backbone.}
Given an input video $\bm{V}$, we first extract video features $\bm{F}\in\mathbb{R}^{L \times D}$ through a backbone network $g$:
\begin{align}
    \bm{F} = g(\bm{V}),
\end{align}
where $D$ denotes the feature dimension. We employ pre-trained ResNet-50~\cite{resnet} as $g$.

\paragraph{Encoder.}
The encoder $f$ is designed to capture diverse temporal variations between adjacent frames by leveraging temporal self-similarity, which helps identify subtle changes in scene dynamics that indicate event boundaries. Following~\cite{zheng2024rethinking}, we adopt the temporal self-similarity encoder as $f$. 
Specifically, the encoder $f$ comprises a 1D convolution (kernel size=3), followed by a sliding-window temporal self-similarity module, a fully convolutional network (FCN), and a 2D pooling operation, as shown in Fig.~\ref{fig:architecture}.
The encoder $f$ takes video features $\bm{F}$ as input and produces temporal embeddings $\bm{E}\in\mathbb{R}^{L \times C}$ as output:
\begin{align}
    \bm{E} = f(\bm{F}),
\end{align}
where $C$ denotes the feature dimension. For further architectural details, please refer to~\cite{zheng2024rethinking}.

\paragraph{Decoder.}
The decoder $h$ is built upon the Transformer encoder layer~\cite{vaswani2017attention}, denoises the input noisy boundary labels $\bm{y}_t$ at time-step $t$ into ground-truth labels, conditioned on temporal embeddings $\bm{E}$.
As illustrated in Fig.~\ref{fig:architecture}, the input $\bm{y}_t$ is first processed by an \texttt{MLP} layer, then concatenated with $\bm{E}$ along the channel dimension. A sinusoidal position embedding is added to this combined feature, which is then fed into self-attention layers~\cite{vaswani2017attention}.
The diffusion time step $t$, encoded via sinusoidal embedding and \texttt{MLP} layers, is injected into the model through a scale-and-shift operation~\cite{ho2020denoising, perez2018film}. Finally, the output from the decoder $h$ is processed by an \texttt{MLP} layer to produce the final prediction $\hat{\bm{y}}_t$:
\begin{align}
\hat{\bm{y}}_t = h(\bm{y}_t, t, \bm{E}).
\label{eq:decoder}
\end{align}

\subsection{Training objective}
\label{sec:training_object}
The model is trained using mean squared error loss $\mathcal{L}$ between the ground-truth boundary label $\bm{y}_0$ and the prediction $\hat{\bm{y}}_t$ at time-step $t$:
\begin{equation}
\mathcal{L} = \frac{1}{L} \sum_{l=1}^{L} (\bm{y}_{0,l} - \hat{\bm{y}}_{t, l})^2.
\end{equation}

\subsection{Classifier-free guidance (CFG)}
\label{sec:cfg}
To address the inherent ambiguity in event boundary detection, we use classifier-free guidance~\cite{ho2021classifier}. This guidance strategy balances prediction diversity and fidelity by combining conditional and unconditional diffusion models. 

\paragraph{Training with CFG.}
Both conditional and unconditional diffusion models are trained for classifier-free guidance. To achieve this, we randomly drop the conditional features $\bm{E}$ with probability $p\in[0, 1]$ in Eq.~\ref{eq:decoder}, jointly training the model to predict with and without conditioning:

\begin{equation}
\hat{\bm{y}}_t = \begin{cases}
\hat{\bm{y}}^{\text{c}}_t = h(\bm{\bm{y}}_t, t, \bm{E}), & \text{with probability } 1 - p, \\
\hat{\bm{y}}^{\text{u}}_t = h(\bm{\bm{y}}_t, t, \bm{0}_{L \times C}), & \text{with probability } p.
\end{cases}
\label{eq:cfg_train}
\end{equation}
where $\bm{0}_{m \times n}$ denotes a zero matrix with size of $m$ and $n$.

\paragraph{Inference with CFG.}
During inference, diversity can be adjusted by changing the value of classifier-free guidance weight $w$:
\begin{equation}
\hat{\bm{y}}_t = (1 + w)\hat{\bm{y}}_t^\mathrm{c} - w\hat{\bm{y}}_t^\mathrm{u},
\label{eq:cfg_inf}
\end{equation}
where $\hat{\bm{y}}_t^\mathrm{c}$ and $\hat{\bm{y}}_t^\mathrm{u}$ denote the conditional and unconditional predictions, respectively, obtained from Eq.~\ref{eq:cfg_train}.
A larger $w$ leads to more deterministic predictions that closely follow the video content, while a smaller $w$ allows for more diverse predictions that reflect the inherent ambiguity in boundary distribution.
The overall training and inference algorithms are provided in the supplementary material.
\section{Experiments}
\label{sec:experiments}

\subsection{Setup}
\label{sec:datasets}
In our experiment, we evaluate our method on two standard GEBD benchmarks: Kinetics-GEBD~\cite{shou2021generic} and TAPOS~\cite{shao2020intra}. 
Each video is uniformly sampled to 100 frames. We use ResNet-50~\cite{resnet} pretrained on ImageNet-1K~\cite{Imagenet} as the backbone network $g$.
We employ the BasicGEBD-L4 encoder~\cite{zheng2024rethinking} and a 6-layer Transformer~\cite{vaswani2017attention} for our encoder $f$ and decoder $h$, . We adopt FiLM~\cite{perez2018film} for the diffusion timestep embedding. 
During training, we set probability $p$ of classifier-free guidance as 0.1.
For Kinetics-GEBD~\cite{shou2021generic}, which provides five annotations per video, we use a maximum of four annotations, selected based on F1 consistency score~\cite{shou2021generic, li2022structured, zheng2024rethinking}.
Please refer to our supplementary materials for more details of the datasets and our implementation.

\subsection{Evaluation Metrics}
\label{sec:evaluation_protocol}
\subsubsection{F1 score}
\label{sec:evaluation_protocol_convention}
In the conventional evaluation of GEBD, a single prediction is evaluated for each video~\cite{shou2021generic, zheng2024rethinking, li2022structured, zheng2025fine}.
The F1 score based on relative distance (Rel.Dis.~\cite{shou2021generic}) is the basic evaluation metric.
When multiple annotations are available, the F1 score is computed by taking the maximum F1 score among all possible prediction-annotation pairs.

However, the F1 score does not account for scenarios where multiple solutions are generated, nor does it capture the inherent diversity among ground-truth annotations.
In the following, we introduce new evaluation metrics, \ie, \textit{symmetric F1} and \textit{diversity} scores, that consider both multiple predictions and the diversity of GT annotations.

\subsubsection{Symmetric F1 score}
\label{sec:evaluation_protocol_diversity}
When multiple predictions are generated for a video, evaluating the many-to-many alignment between predictions and GT annotations requires considering two key aspects: (1) how accurately each prediction matches one of the GT annotations (Pred-to-GT alignment) and (2) how well each GT annotation is covered by the predictions (GT-to-Pred alignment).
To address these aspects, we propose the symmetric F1 score (\text{F1}$_{\text{sym}}$), which combines two directional F1 scores: the Pred-to-GT alignment score (\text{F1}$_{\text{p2g}}$) and the GT-to-Pred alignment scores (\text{F1}$_{\text{g2p}}$). This bi-directional metric ensures a comprehensive evaluation by jointly measuring how well predictions capture the ground truth and vice versa, reflecting both prediction accuracy and diversity.

To formally define our metrics, we first establish our notation.
For each video with $L$ frames, we denote $N_\mathrm{G}$ ground truth annotations and $N_\mathrm{P}$ model predictions by $\bm{Y}\in\mathbb{R}^{N_\mathrm{G}\times L}$ and $\hat{\bm{Y}}\in\mathbb{R}^{N_\mathrm{P}\times L}$, respectively. 

\smallskip
\noindent \textbf{Pred-to-GT alignment score $\text{F1}_\text{p2g}$.}
The Pred-to-GT alignment score, $\text{F1}_\text{p2g}$, measures how well each predicted boundary aligns with at least one ground truth annotation, similar to the conventional GEBD evaluation.
It is computed by taking each prediction $\hat{Y}_i$, finding its highest F1 score across all ground truth annotations $Y_j$, and averaging these maximum scores across all predictions as:
\begin{equation}
    \text{F1}_{\text{p2g}}(\hat{\bm{Y}}, \bm{Y}) = \frac{1}{N_\mathrm{P}} \sum_{i=1}^{N_\mathrm{P}} \max_{j\in\{1,\ldots,N_\mathrm{G}\}} \text{F1}(\hat{Y}_i, Y_j),
    \label{eq:F1_p2g}
\end{equation}
where $\text{F1}(X, Y)$ computes the F1 score between $X$ and $Y$.

\smallskip
\noindent \textbf{GT-to-Pred alignment score $\text{F1}_\text{g2p}$.}
To account for the variability and diversity in GT annotations, the GT-to-Pred score, $\text{F1}_{\text{g2p}}$, evaluates how well each annotation is covered by any of the predictions.  
This is achieved by reversing the formulation to assess each ground truth annotation against all predictions, as follows:
\begin{equation}
    \text{F1}_{\text{g2p}}(\hat{\bm{Y}}, \bm{Y}) = \frac{1}{N_\mathrm{G}} \sum_{j=1}^{N_\mathrm{G}} \max_{i\in\{1,\ldots,N_\mathrm{P}\}} \text{F1}(\hat{Y}_i, Y_j).
\end{equation}

\smallskip
\noindent \textbf{Symmetric F1 score $\text{F1}_\text{sym}$.}
The symmetric F1 score finally combines the two directional F1 scores, \ie, $\text{F1}_{\text{p2g}}$ and $\text{F1}_{\text{g2p}}$, by taking a harmonic mean as:
\begin{equation}
    \text{F1}_{\text{sym}}(\hat{\bm{Y}}, \bm{Y}) = \frac{2\times \text{F1}_{\text{p2g}}(\hat{\bm{Y}}, \bm{Y}) \times \text{F1}_{\text{g2p}}(\hat{\bm{Y}}, \bm{Y})}{\text{F1}_{\text{p2g}}(\hat{\bm{Y}}, \bm{Y}) + \text{F1}_{\text{g2p}}(\hat{\bm{Y}}, \bm{Y})}.
\end{equation}
The final symmetric F1 score for the entire dataset is obtained by computing the score for each video individually and then taking an average across all videos in the dataset.

\subsubsection{Diversity score}
Although the proposed symmetric F1 score measures a comprehensive alignment between multiple predictions and ground truth annotations, it does not directly measure the diversity among predictions.
We thus introduce the diversity score that directly quantifies the average pairwise dissimilarity among predictions, following~\cite{10.1145/3503161.3548115}.
The diversity score among $N_\mathrm{P}$ predictions $\hat{\bm{Y}}$ is defined as:

\begin{equation}
    \text{Diversity}(\hat{\bm{Y}}) = \frac{1}{N_\mathrm{P}^2} \sum_{i=1}^{N_\mathrm{P}} \sum_{j=1}^{N_\mathrm{P}} (1 - \text{F1}(\hat{Y}_i, \hat{Y}_j)),
\end{equation}

which computes the average dissimilarity among all predictions. 
Here, the F1 score serves as the similarity measure, ensuring that the diversity score reflects how different the generated predictions are from one another. 
Note that higher values indicate greater diversity.
Similar to the symmetric F1 score, the diversity score is averaged across all videos in the dataset.

\begin{figure}[t!]
     \centering
     \resizebox{0.9\linewidth}{!}{
     \includegraphics[width=\linewidth]{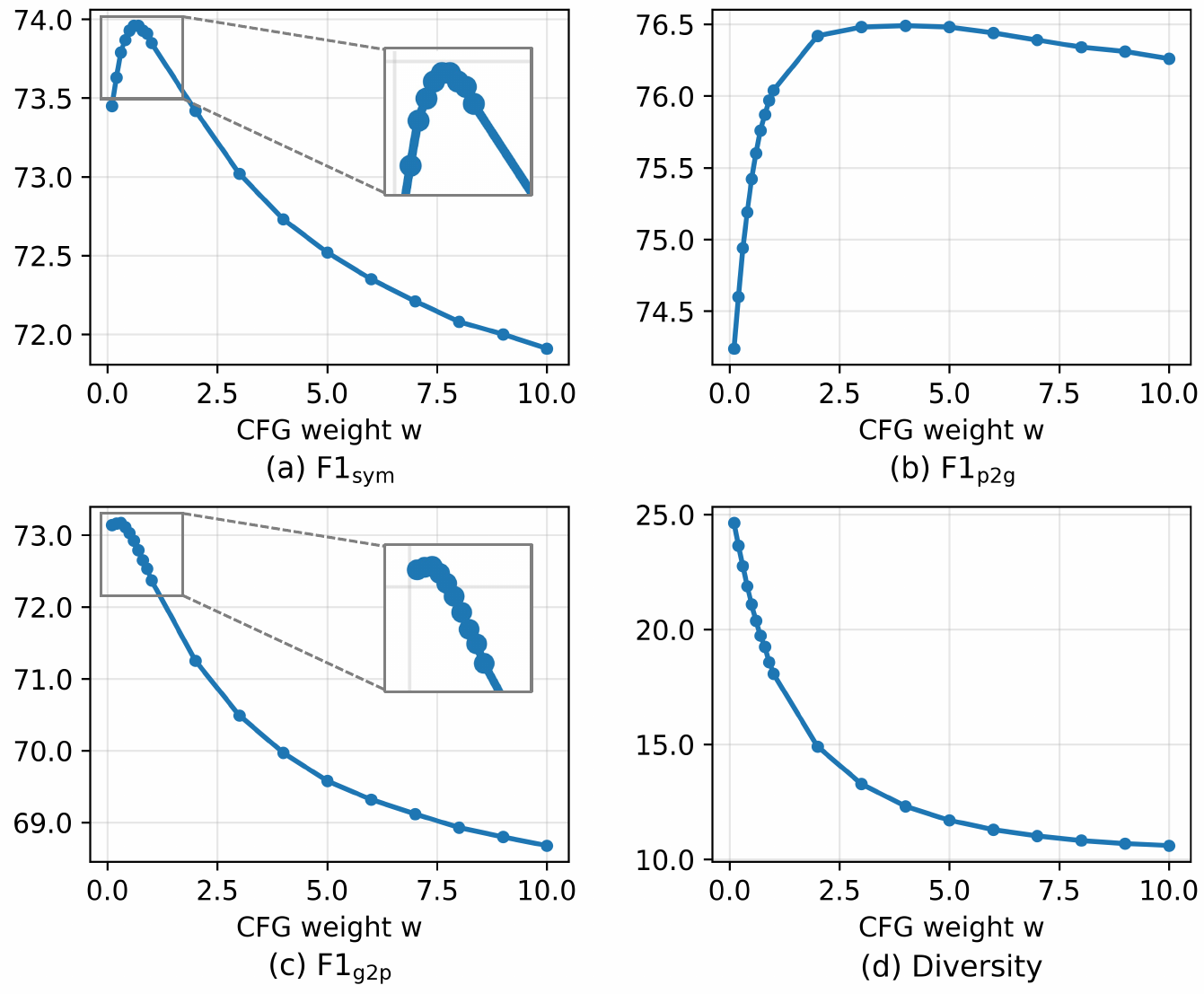}
     } 
     \vspace{-3mm}
     \caption{\textbf{Effect of CFG weight $w$.} The x-axis represents the CFG weight $w$, while the y-axis shows (a) $\text{F1}_{\text{sym}}$, (b) $\text{F1}_{\text{p2g}}$, (c)$\text{F1}_{\text{g2p}}$, and (d) diversity, respectively.}
     
     \label{fig:guidance}
     \vspace{-6mm}
\end{figure}

\subsection{Effect of the CFG Weight $w$}

The CFG weight $w$ is a key factor in balancing the conditional and unconditional diffusion models. A larger $w$ increases the influence of the conditional model, strengthening the impact of the temporal self-similarity feature in the diffusion process. In contrast, a smaller $w$ increases the influence of the unconditional model, enabling the generation of more diverse predictions by relying less on the conditioning signal.
To evaluate the effect of $w$ in Eq.~\ref{eq:cfg_inf}, we vary its value from $0.0$ to $10.0$ during inference, following~\cite{ho2021classifier}.
Figure~\ref{fig:guidance} presents the results, where the x-axis corresponds to the value of $w$, while the y-axis indicates (a) $\text{F1}_\text{sym}$, (b) $\text{F1}_\text{p2g}$, (c) $\text{F1}_\text{g2p}$, and (d) diversity score.

The Pred-to-GT alignment score $\text{F1}_{\text{p2g}}$ initially increases with larger values of $w$ (Fig.~\ref{fig:guidance}-(b)), as the model benefits from stronger conditioning of temporal self-similarity. However, when $w$ becomes greater than 5.0, the score starts to decline, likely due to over-reliance on the conditional signal, which may hinder the model from capturing subtle motion.
In contrast, the GT-to-Pred alignment score $\text{F1}_\text{g2p}$ (Fig.\ref{fig:guidance}-(c)) and the diversity score (Fig.\ref{fig:guidance}-(d)) exhibit a similar trend—both decrease as $w$ increases, since stronger conditioning reduces variability in the generated predictions. 
However, when $w$ becomes too small (i.e., close to zero), $\text{F1}_\text{g2p}$ also drops, as overly diverse samples tend to deviate from the ground truth, making alignment more difficult.
Overall results suggest that higher $\text{F1}_{\text{p2g}}$ score does not always guarantee diverse predictions, and that excessive diversity may negatively impact Pred-to-GT alignment.

These observations motivate the introduction of a unified metric that captures both diversity and fidelity of the generated predictions.
The symmetric F1 score $\text{F1}_\text{sym}$, defined as the harmonic mean of $\text{F1}_{\text{p2g}}$ and $\text{F1}_{\text{g2p}}$, exhibits a non-monotonic relationship with the guidance weight, reaching its peak at $w = 0.6$. 
This result highlights the trade-off between Pred-to-GT alignment and GT-to-Pred alignment.
A moderate guidance weight effectively balances these trade-offs, maximizing the symmetric F1 score by preserving alignment with the ground truth while ensuring sufficient diversity in predictions.
The complete numerical results are provided in the supplementary material.
\begin{table}[t]
    \centering
    \resizebox{\linewidth}{!}{
    \begin{tabular}{lcccc}
        \toprule
        Method & $\text{F1}_{\text{sym}}$ & $\text{F1}_{\text{p2g}}$ & $\text{F1}_{\text{g2p}}$ & Diversity \\
        \midrule
        Temporal Perceiver$^\dagger$~\cite{tan2023temporal} & 69.4 & 72.2 & 67.4 & 14.6 \\
        SC-Transformer$^\dagger$~\cite{li2022structured} & \underline{72.9} & 74.9 & \underline{71.6} & \underline{18.9} \\
        BasicGEBD$^\dagger$~\cite{zheng2024rethinking} & 72.2 & 74.5 & 70.6 & 18.6 \\
        EfficientGEBD$^\dagger$~\cite{zheng2024rethinking} & 72.6 & \textbf{76.0} & 70.2 & 14.9 \\
        \rcol \textbf{\ours~(ours)} & \textbf{74.0} & \underline{75.6} & \textbf{72.9} & \textbf{20.4} \\
        \bottomrule
    \end{tabular}
    }
    \vspace{-3mm}
    \caption{\textbf{Diversity-aware evaluation on Kinetics-GEBD.} \textbf{Boldface} and \underline{underline} indicate the best and the second-best scores. $^\dagger$ Results are obtained using reproduced models.}
    \label{tab:diversegebd}
    \vspace{-6mm}
\end{table}

\subsection{Diversity-aware Evaluation of GEBD}
\label{sec:diversegebd}
In Table~\ref{tab:diversegebd}, we compare \ours~with previous methods~\cite{tan2023temporal, li2022structured, zheng2024rethinking} on the Kinetics-GEBD dataset.
For multiple prediction generations, we set the number of predictions $N_\mathrm{P}$ to 5, as the average number of annotations per video in the dataset is 4.93~\cite{shou2021generic}.
Since all previous methods produce deterministic outputs, we reproduce and evaluate each model by training it five times with random initialization to obtain multiple predictions. Please note that our experiments are conducted on models with publicly available code\footnote{We utilize the official Github repositories for Temporal Perceiver~\cite{tan2023temporal}: \url{https://github.com/MCG-NJU/TemporalPerceiver}, SC-Transformer~\cite{li2022structured}: \url{https://github.com/lufficc/SC-Transformer}, and BasicGEBD/EfficientGEBD~\cite{zheng2024rethinking}: \url{https://github.com/Ziwei-Zheng/EfficientGEBD}.}.
The reproduced models are marked with $^\dagger$, and their performances is reported in the supplementary material.
Unlike these deterministic models, \ours~generates diverse predictions from a single trained model by varying the initial Gaussian noise $\hat{\bm{y}}$, eliminating the need for multiple training runs to achieve diversity. In this experiment, we set the CFG weight $w$ to 0.6 and the relative distance threshold for the F1 score to 0.05.

Table~\ref{tab:diversegebd} presents the overall results, where \ours~achieves the state-of-the-art performance on $\text{F1}_\text{sym}$, $\text{F1}_\text{g2p}$, and the diversity score, while showing comparable results on $\text{F1}_\text{p2g}$ compared to the previous methods.
These results indicate that \ours~is capable of generating diverse predictions while maintaining strong alignment with ground-truth annotations, thereby achieving an effective balance between diversity and plausibility.
EfficientGEBD~\cite{zheng2024rethinking} achieves the highest score in $\text{F1}_\text{p2g}$; however, its lower $\text{F1}_\text{g2p}$ results in a reduced $\text{F1}_\text{sym}$, and its diversity score is also notably low.
These results suggest that the model generates highly precise but less diverse predictions, covering fewer ground-truth annotations and prioritizing precision over diversity.
By comparing the results of EfficentGEBD to BasicGEBD~\cite{zheng2024rethinking}, we observe that a significant increase in diversity does not necessarily lead to a proportional improvement in $\text{F1}_\text{g2p}$. 
This finding implies that higher diversity alone does not guarantee better GT-to-Pred alignment, emphasizing the importance of plausibility in predictions. 
Full results with varying relative distance values are presented in the supplementary material.

\subsection{Analysis}
\label{sec:analysis}
\begin{figure}[]
     \centering
     \resizebox{0.9\linewidth}{!}{
     \includegraphics[width=\linewidth]{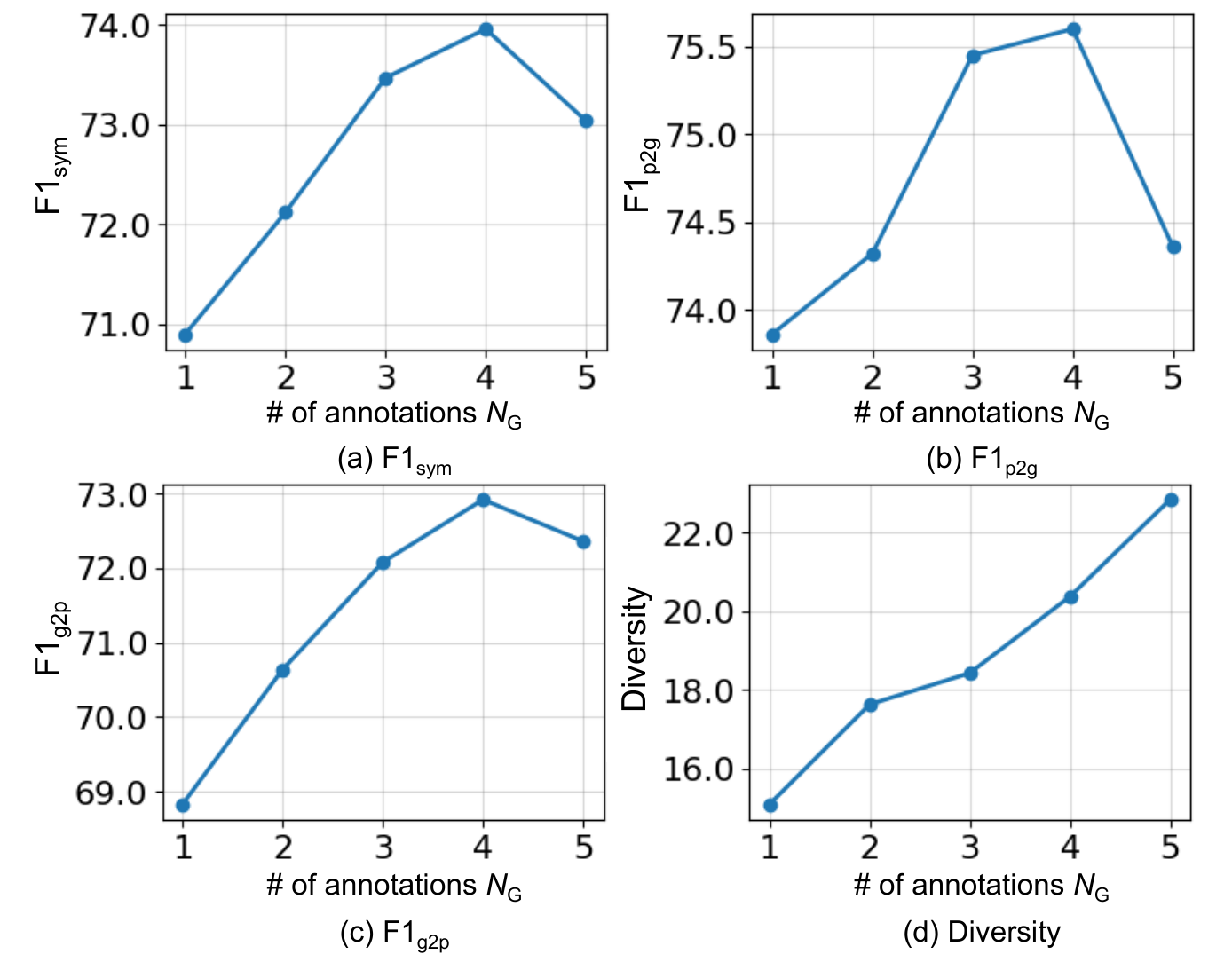}
     } 
     \vspace{-3mm}
     \caption{\textbf{Effect of the number of annotations.} Model performance with varying numbers of annotations (1-5).}
     \label{fig:annotator}
     \vspace{-3mm}
\end{figure}

\begin{table}[t]
    \centering
    \resizebox{\linewidth}{!}{
    \begin{tabular}{C{2cm}C{1.5cm}C{1.5cm}C{1.5cm}C{1.5cm}}
    \toprule
    Conditioning&$\text{F1}_{\text{sym}}$&$\text{F1}_{\text{p2g}}$&$\text{F1}_{\text{g2p}}$& Diversity \\
    \midrule
    $\bm{F}$ & 68.5 & 69.3 & 68.3 & \textbf{24.3} \\ 
    \rcol $\bm{E}$ & \textbf{74.0} & \textbf{75.6} & \textbf{72.9} & 20.4 \\
    \bottomrule
    \end{tabular}
    }
    \vspace{-3mm}
    \caption{\textbf{Effect of conditioning in diffusion.} Using temporal self-similarity feature $\bm{E}$ as a diffusion condition is effective.}
    \label{tab:ans_condition}
    \vspace{-6mm}
\end{table}
\paragraph{Effect of the number of annotations $N_\mathrm{G}$.}
Since each annotation represents individual subjective interpretations of event boundaries, we experiment by adjusting the number of annotations $N_\mathrm{G}$ used during training. 
Instead of random selection, we prioritize annotations based on their reliability measured by the F1 consistency scores~\cite{shou2021generic}. Specifically, we increase $N_\mathrm{G}$ from 1 to 5 by selecting the top-$N_\mathrm{G}$ annotations with the highest consistency scores.

Figure~\ref{fig:annotator} presents the results, where the x-axis denotes $N_\mathrm{G}$, and the y-axis shows (a) $\text{F1}_\text{sym}$, (b) $\text{F1}_\text{p2g}$, (c) $\text{F1}_\text{g2p}$, and (d) diversity score.
We observe a consistent improvement in overall performance as $N_\mathrm{G}$ increases from 1 to 4, indicating that incorporating multiple reliable annotators helps the model better capture variations in boundary annotations while improving fidelity.
However, when all five annotators are included, we observe a decline in $\text{F1}_\text{sym}$, $\text{F1}_\text{p2g}$, and $\text{F1}_\text{g2p}$, while the diversity score continues to increase.
This suggests that although using more annotations enhances diversity, incorporating low-consistency annotations can negatively impact performance.

\paragraph{Effect of conditioning in diffusion.}
To examine the effect of the conditioning feature in denoising diffusion, we conduct experiments by varying the conditioning feature in the diffusion process. Specifically, we replace the temporal self-similarity feature $\bm{E}$ with visual features $\bm{F}$ extracted directly from the backbone network $g$. Table~\ref{tab:ans_condition} presents the results.
We observe a significant performance drop when using $\bm{F}$, demonstrating the importance of temporal self-similarity features as a conditioning input for the diffusion model. Since self-similarity captures subtle changes across frames, using $\bm{E}$ is more effective. 

\begin{table}[t]
    \centering
    \resizebox{1.0\linewidth}{!}{
    \begin{tabular}{C{1.5cm}C{1.5cm}C{1.5cm}C{1.5cm}C{1.5cm}}
    \toprule   Steps&$\text{F1}_{\text{sym}}$&$\text{F1}_{\text{p2g}}$&$\text{F1}_{\text{g2p}}$& Diversity \\
    \midrule
    1 & 64.0 & 71.3 & 59.1 & 9.2 \\
    2 & 72.3 & 75.2 & 70.8 & 17.9 \\
    4 & 73.4 & 75.5 & 71.9 & 18.5 \\
    8 & 73.7 & \textbf{75.6} & 72.4 & 19.4 \\
    16 & 73.8 & 75.4 & 72.6 & 19.9 \\
    \rcol 32 & \textbf{74.0} & \textbf{75.6} & \textbf{72.9} & 20.4 \\
    50 & 73.9 & 75.5 & \textbf{72.9} & \textbf{20.8} \\
    \bottomrule
    \end{tabular}
    }
    \vspace{-2mm}
    \caption{\textbf{Effect of inference step.} Following the DDIM sampling strategy, the model can skip the timestep T.}
    \vspace{-2mm}
    \label{tab:ans_step}
\end{table}
\begin{table}[t]
    \centering
    \resizebox{0.9\linewidth}{!}{
    \begin{tabular}{l c c}
    \toprule
    \multirow{2}*[-0.5ex]{Method} & \multicolumn{2}{c}{F1@0.05} \\
    \cmidrule[0.3pt]{2-3}
    & Kinetics-GEBD & TAPOS \\
    \midrule
    BMN~\cite{Lin_2019_ICCV}&18.6&-\\
    BMN-StartEnd~\cite{Lin_2019_ICCV}&49.1&-\\
    ISBA~\cite{ding2018weakly}&-&10.6\\
    TCN~\cite{lea2016segmental}&58.8&23.7\\
    CTM~\cite{huang2016connectionist}&-&24.4\\
    TransParser~\cite{shao2020intra}&-&23.9\\
    PC~\cite{shou2021generic}&62.5&52.2\\
    SBoCo~\cite{kang2022uboco}&73.2&-\\
    Temporal Perceiver~\cite{tan2023temporal}&74.8&55.2\\
    DDM-Net~\cite{tang2022progressive}&76.4&60.4\\
    CVRL~\cite{li2022end}&74.3&-\\
    LCVS~\cite{zhang2024local}&76.8&-\\
    SC-Transformer~\cite{li2022structured}&77.7&61.8\\
    BasicGEBD~\cite{zheng2024rethinking}&76.8&60.0\\
    EfficientGEBD~\cite{zheng2024rethinking}&78.3&\underline{63.1}\\
    DyBDet~\cite{zheng2025fine}&\textbf{79.6}&62.5\\
    \midrule
    \rcol \textbf{DiffGEBD (ours)}&\underline{78.4}&\textbf{65.8}\\
    \bottomrule
    \end{tabular}
    }
    \vspace{-2mm}
    \caption{
    \textbf{Conventional evaluation of GEBD}
    }
    \label{tab:deterministic}
    \vspace{-4mm}
\end{table}

\begin{figure*}[!t]
     \centering
     \resizebox{\linewidth}{!}{
     \includegraphics[width=0.9\textwidth]{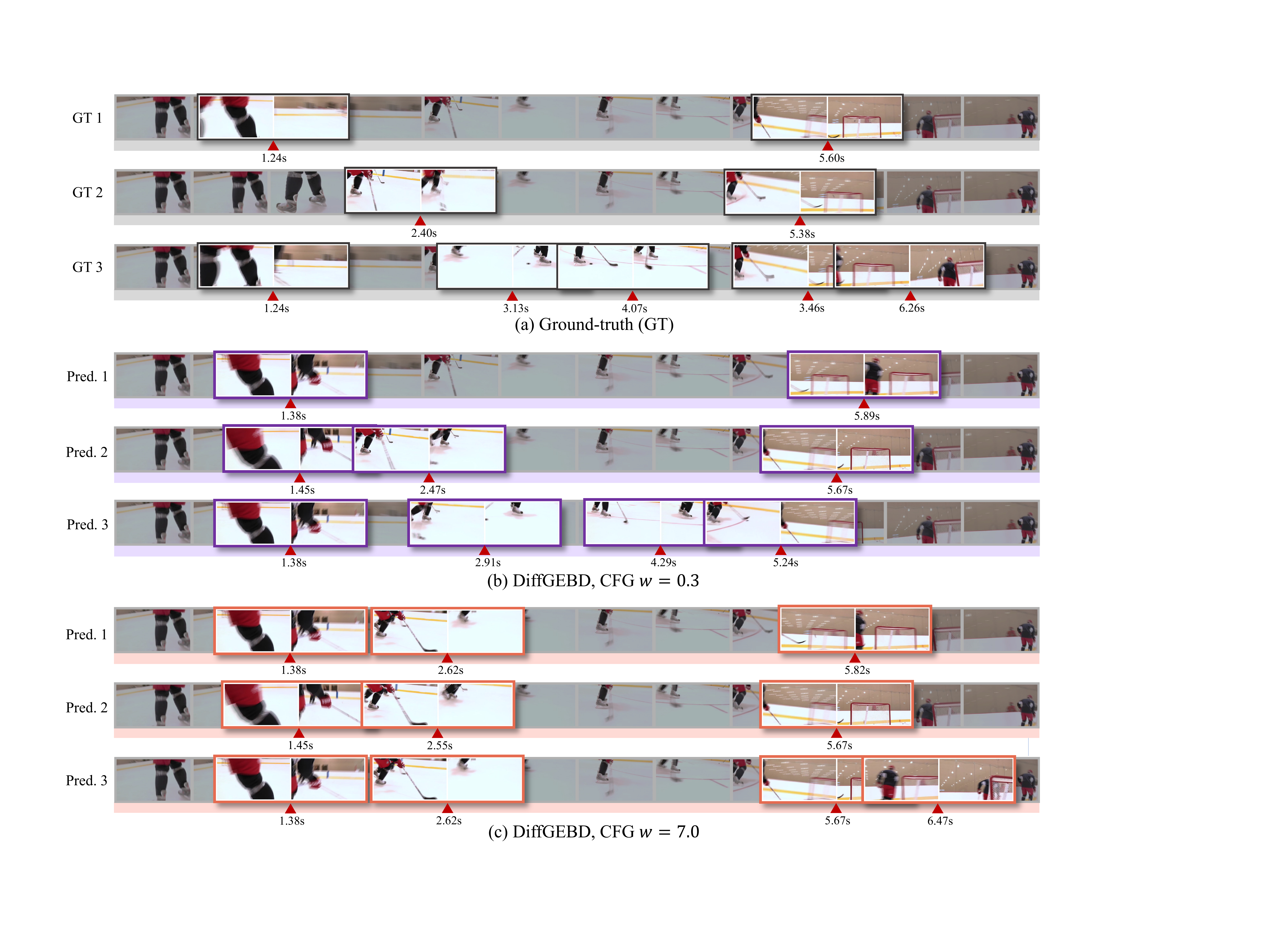}
     } 
     \vspace{-7mm}
     \caption{\textbf{Example results on Kinetics-GEBD.}
     The figure illustrates (a) Ground-truth annotations, (b) predictions with $w = 0.3$, and (c) predictions with $w = 7.0$.
     }
     \label{fig:qual}
     \vspace{-6mm}
\end{figure*}

\paragraph{Effect of the number of inference steps.}
We investigate the impact of the number of diffusion inference steps by varying T from 1 to 50. 
As shown in Table~\ref{tab:ans_step}, overall performance improves as T increases, but no further gains are observed when T exceeds 32. Therefore, we set T to 32 as the optimal number of steps.

\subsection{Conventional Evaluation of GEBD}
\label{sec:detegebd}
Table~\ref{tab:deterministic} compares the performance of the proposed method on two standard GEBD benchmark datasets, Kinetics-GEBD and TAPOS, under the conventional evaluation setting, following previous method~\cite{shou2021generic}.
Note that all methods use ResNet-50~\cite{he2016deep} trained on ImageNet~\cite{Imagenet} as the backbone network for a fair comparison.
In this experiment, we set the CFG weight $w$ to 4.0 to improve the fidelity of the predictions by strengthening the influence of temporal self-similarity features in the video.
\ours~achieves comparable results on Kinetics-GEBD and outperforms prior methods on TAPOS.
These results demonstrate that \ours~ can effectively generate highly feasible predictions with a high guidance weight, ensuring stronger adherence to the conditioning features.

\subsection{Example Results}
\label{sec:qual}

Figure~\ref{fig:qual} illustrates example results of \ours~on the Kinetics-GEBD dataset, showing (a) ground-truth annotations, (b) predictions with $w=0.3$, and (c) predictions with $w=7.0$.
All outputs were generated using the same model with different initial noise.
We observe that clear boundaries (e.g., subject's movements between 1.24s to 1.45s) are consistently detected across the predictions, regardless of the guidance weight. However, boundaries that exhibit human ambiguity, such as subtle action changes (e.g., hockey stick movements at 2.91s and 4.29s in Pred. 3 of (b)), vary across different generations. Notably, we observe that lower weight guidance allow for diverse predictions, while higher guidance weights lead to more consistent predictions.
\vspace{-3mm}
\section{Conclusion}
\label{sec:conclusion}

We have presented \ours, a diffusion-based boundary detection model from a generative perspective. The proposed method encodes temporal dynamics based on self-similarity, then iteratively refines the Gaussian noise into plausible boundaries via denoising diffusion.
By integrating classifier-free guidance, our model enables to explicitly control the degree of diversity. Furthermore, we have introduced the symmetric F1 and diversity scores, which jointly capture many-to-many alignments and the variability in model predictions. We believe that our model offers a new perspective on producing diverse yet 
\section*{Acknowledgement}
\label{sec:acknowledgement}

This work was supported by Samsung Electronics (IO201208-07822-01), the NRF research grant (RS-2021-NR059830 (40\%)), the IITP grants (RS-2022-II220264: Comprehensive Video Understanding and Generation (35\%), RS-2019-II191906: AI Graduate School at POSTECH (5\%)) funded by the Ministry of Science and ICT, Korea, and the Scaleup TIPS grant (RS-2023-00321784: Development of Novel Generative AI Technology to Generate Domain-Specific Synthetic Data (20\%)) funded by the Ministry of SMEs and Startups, Korea. 

{
    \small
    \bibliographystyle{ieeenat_fullname}
    \bibliography{main}
}

\renewcommand\thefigure{S\arabic{figure}}
\renewcommand{\thetable}{S\arabic{table}}
\setcounter{figure}{0}
\setcounter{table}{0}
\clearpage
\twocolumn[
    \begin{center}
    \Large\textbf{Generic Event Boundary Detection via Denoising Diffusion \\
    \textit{- Supplementary Material -}}
    \vspace{0.5cm}
    \end{center}
]
\begin{algorithm}[t]
    \small
    \caption{\small \ours~training algorithm}
    \label{alg:train}
    \algcomment{\fontsize{7.2pt}{0em}\selectfont \texttt{alpha\_cumprod(t)}: cumulative product of $\alpha_i$, \ie, $\prod_{i=1}^t \alpha_i$}
    \definecolor{codeblue}{rgb}{0.25,0.5,0.5}
    \definecolor{codekw}{rgb}{0.85, 0.18, 0.50}
    \definecolor{codeblue2}{RGB}{59,71,237}
    \lstset{
      backgroundcolor=\color{white},
      basicstyle=\fontsize{7.2pt}{7.2pt}\ttfamily\selectfont,
      columns=fullflexible,
      breaklines=true,
      captionpos=b,
      commentstyle=\fontsize{7.2pt}{7.2pt}\color{codeblue},
      keywordstyle=\fontsize{7.2pt}{7.2pt}\color{codekw},
      escapechar={|}, 
    }
    \begin{lstlisting}[language=python]
def train_loss(V, T, y_0, p):
    """
    V: video [B, T, H, W, 3]
    T: diffusion time-step
    y_0: ground-truth boundary labels [B, L, 1]
    p: CFG probability
    """

    # Extract features from backbone network g
    F = g(V)

    # Extract visual embeddings from the encoder f
    E = f(F)
    
    # Random sample for time-step
    t = uniform(0, T)

    eps = normal(mean=0, std=1)
    
    # Corrupt data
    y_crpt = sqrt(|~~~|alpha_cumprod(t)) * y_0 + 
              |~|sqrt(1 - alpha_cumprod(t)) * eps
              
    # Classifier-free Guidance by probability p
    |\color{codeblue2}if uniform(0, 1) < p:|
        |\color{codeblue2}E = zeros\_like(E)|

    # Predict with the decoder h
    y_hat = h(y_0, E, t)

    # Mean squared loss
    loss = (y_0 - y_hat)**2
    loss = mean(loss)
    
    return loss
    \end{lstlisting}
\end{algorithm}

In this supplementary material, we present detailed explanations and additional experimental results. Specifically, we include the training and inference algorithms in 
Sec.~\ref{sec_sup:algorithms}, experimental details in Sec.~\ref{sec_sup:experimental_details}, additional experimental results in Sec.~\ref{sec_sup:additional_results}, more example results in Sec.~\ref{sec_sup:more_example}, and a discussion in Sec.~\ref{sec_sup:discussion}.

\section{Algorithms}
\label{sec_sup:algorithms}
We present the training and inference algorithms in Alg.~\ref{alg:train} and Alg.~\ref{alg:inference}, respectively. During training, both conditional and unconditional models are jointly trained with probability $p$, enabling classifier-free guidance. During inference, we iteratively refine the output by balancing the unconditional and conditional outputs according to the guidance weight $w$, obtaining the final output.

\begin{algorithm}[t]
\small
\caption{\small \ours~inference algorithm}
\label{alg:inference}
\definecolor{codeblue}{rgb}{0.25,0.5,0.5}
\definecolor{codekw}{rgb}{0.85, 0.18, 0.50}
\definecolor{codeblue2}{RGB}{59,71,237}
\lstset{
  backgroundcolor=\color{white},
  basicstyle=\fontsize{7.5pt}{7.5pt}\ttfamily\selectfont,
  columns=fullflexible,
  breaklines=true,
  captionpos=b,
  commentstyle=\fontsize{7.5pt}{7.5pt}\color{codeblue},
  keywordstyle=\fontsize{7.5pt}{7.5pt}\color{codekw},
  escapechar={|}, 
}

\begin{lstlisting}[language=python]
def inference(V, T, steps, w):
    """
    V: video [B, T, H, W, 3]
    T: diffusion time step
    steps: the number of inference steps
    w: classifier-free guidance weight
    """
    
    # Extract features from backbone network g
    F = g(V)

    # Extract visual embeddings from the encoder f
    E = f(F)

    y_t = normal(mean=0, std=1)
    
    # Uniform sample step size
    times = reversed(linespace(-1, T, steps))
    time_pairs = list(zip(times[:-1], times[1:]))
    
    for t_now, t_next in zip(time_pairs):
        # conditional prediction
        |\color{codeblue2}y\_hat\_c = h(y\_t, E, t\_now)| 
        
        # unconditional prediction
        |\color{codeblue2}y\_hat\_u = h(y\_t, zeros\_like(E), t\_now)| 
        
        # Form the classifier-free guided prediction
        |\color{codeblue2}y\_hat = (1 + w) * y\_hat\_c - w * y\_hat\_u|

        # Estimate x at t_next
        y_t = ddim_step(y_t, y_hat, t_now, t_next)
    
    return y_t
\end{lstlisting}
\end{algorithm}

\section{Experimental Details}
\label{sec_sup:experimental_details}
\subsection{Datasets}
\paragraph{Kinetics-GEBD.}
Kinetics-GEBD~\cite{shou2021generic} is the largest GEBD dataset, encompassing a wide spectrum of videos. Each boundary is composed of various taxonomy-free boundaries, including action and object changes. 
The dataset includes multiple annotators, with each annotation providing subjective event boundaries.
Each of the training and validation set contains 20K videos from Kinetics-400~\cite{kay2017kinetics}. In our experiments, we report the results on the validation set.

\paragraph{TAPOS.} 
The TAPOS dataset~\cite{shao2020intra} comprises 21 distinct action categories derived from Olympic sports videos. It consists of 13,094 action instances in the training set and 1,790 instances in the validation set. Each video is annotated with a single annotator, which divides a single action into multiple sub-actions. Following~\cite{shou2021generic}, we adapt TAPOS for our GEBD task by trimming each action instance with its action label hidden and conducting experiments on them.

\subsection{Implementation details}

We train our model using AdamW with a batch size of 2 and a learning rate of 2e-5 in all experiments. In determining the final boundary predictions, we identify consecutive predictions that exceed a predefined threshold $\delta$ as boundary candidates. The midpoint of each boundary candidate sequence is then designated as the final boundary prediction~\cite{li2022structured, zheng2024rethinking}.
We set $\delta$ to 0.5 for both Kinetics-GEBD~\cite{shou2021generic} and 0.3 for TAPOS~\cite{shao2020intra} in our experiments.

\section{Additional Experimental Results}
\label{sec_sup:additional_results}
We present additional experimental results following the same settings as in the main paper. All experiments were conducted on the Kinetics-GEBD dataset.

\paragraph{Effect of diffusion process.} We evaluate our diffusion-based approach against two primary baselines at Table~\ref{tab:eff_diff}: a non-diffusion method training multiple times, and a CVAE-based~\cite{NIPS2015_8d55a249} model. Our method outperforms these alternatives, as diffusion models are inherently capable of accurately approximating complex data distributions, which led to our superior performance.
\begin{table}[t]
    \centering
    \resizebox{1.0\linewidth}{!}{%
    \begin{tabular}{lccccccc}
    \toprule
    \multirow{2}{*}{Model} &\multirow{2}{*}{Diffusion} & \multicolumn{4}{c}{Diversity-aware GEBD} & \multicolumn{1}{c}{Conventional GEBD} \\
    \cmidrule(lr){3-6} \cmidrule(lr){7-7}
    && $\mathrm{F1}_{\mathrm{sym}}$ & $\mathrm{F1}_{\mathrm{p2g}}$ & $\mathrm{F1}_{\mathrm{g2p}}$ & Diversity & F1@0.05 \\
    \midrule
    cVAE & - & 62.7 & 66.8 & 59.9 & 15.2 & 70.0 \\
    \ours & - & 73.4 & 75.2 & 72.3 & 20.2 & 77.5 \\
    \rcol \ours & \checkmark & \textbf{74.0} & \textbf{75.6} & \textbf{72.9} & \textbf{20.4} & \textbf{78.4} \\
    \bottomrule
    \end{tabular}%
    }
    \caption{\textbf{Effects of the diffusion process.} The diffusion-based approach consistently outperforms baselines across all metrics.}
    \label{tab:eff_diff}
\end{table}

\paragraph{Effect of the different samplers.} We adopt DPM-Solver++~\cite{lu2025dpm} and UniPC~\cite{zhao2023unipc} samplers for diffusion inference. As in Table~\ref{tab:solver}, the performance remains consistent across different samplers, showing its generalizability.
\begin{table}[t]
    \centering
    \resizebox{1.0\linewidth}{!}{%
    \begin{tabular}{lccccccc}
    \toprule
    \multirow{2}{*}{Model} &\multirow{2}{*}{Sampler} & \multicolumn{4}{c}{Diversity-aware GEBD} & \multicolumn{1}{c}{Conventional GEBD} \\
    \cmidrule(lr){3-6} \cmidrule(lr){7-7}
    && $\mathrm{F1}_{\mathrm{sym}}$ & $\mathrm{F1}_{\mathrm{p2g}}$ & $\mathrm{F1}_{\mathrm{g2p}}$ & Diversity & F1@0.05 \\
    \midrule
    \ours & DPM-Solver++ & 73.8 & \textbf{76.0} & 72.2 & 18.0 & 78.2 \\
    \ours & UniPC & 73.8 & 76.1 & 72.2 & 17.7 & 78.4 \\
    \rcol \ours & DDIM & \textbf{74.0} & 75.6 & \textbf{72.9} & \textbf{20.4} & \textbf{78.4} \\
    \bottomrule
    \end{tabular}%
    }
    \caption{\textbf{Effect of the diffusion sampler.}}
    \label{tab:solver}
\end{table}
\begin{table}[t]
\centering
\resizebox{\linewidth}{!}{
\begin{tabular}{lccccccc}
\toprule
Model & F1$_{\mathrm{sym}}$ &Div. &Train time & Inf. time & Mem. & \#param\\
\midrule
Temporal Perciever~\cite{tan2023temporal} & 69.4 & 14.6 &  8.7h  &  \textbf{0.03s}  &  \textbf{0.1G}  &  52.2M \\
SC-Transformer~\cite{li2022structured}    & 72.9 & 18.9 & 44.7h & 0.15s & 9.4G & 71.6M \\
BasicGEBD~\cite{zheng2024rethinking}        & 72.2 & 18.6 & 19.9h & 0.15s & 7.2G & \textbf{32.2M} \\
EfficientGEBD~\cite{zheng2024rethinking}     & 72.6 & 14.9 & 16.4h & 0.15s & 6.2G & 33.2M \\
\midrule
\rcol \ours~(4 steps)      & 73.4 & 18.5 & 6.6h  & 0.16s & 7.1G & 68.0M \\
\rcol \ours~(8 steps)      & 73.7 & 19.4 & 6.6h  & 0.19s & 7.1G & 68.0M \\
\rcol \ours~(32 steps)     & \textbf{74.0} & \textbf{20.4} & \textbf{6.6h}  & 0.36s & 7.1G & 68.0M \\
\bottomrule
\end{tabular}
} 
\vspace{-2mm}
\centering
\caption{\textbf{Computational cost on Diversity-aware evaluation.} We report the training time and parameters per whole video and the inference time and memory per frame.}
\vspace{-4mm}
\label{tab:computational_efficiency}
\end{table}

\paragraph{Computational cost on the diversity-aware evaluation.}
Table~\ref{tab:computational_efficiency} compares the computational efficiency of our method with others reported in Table~\ref{tab:diversegebd}.
All results are obtained using RTX 6000 Ada GPU under the same settings. For deterministic methods, we report the total training time across five runs. Temporal Perciever~\cite{tan2023temporal} uses pre-extracted features without end-to-end backbone fine-tuning, which explains its lower Inf. time and memory usage.
Compared to EfficientGEBD, our method achieves substantially lower training time and comparable inference time with 4 sampling steps, while still outperforming it.
Although inference time increases with more steps, it can be reduced via recent advances, \eg, Flow Matching~\cite{lipmanflow}, which we leave for future work. In terms of memory footprint and model size, our method maintains similar memory usage to other baselines using a moderate number of parameters, offering a good balance between efficiency and capacity.

\begin{table}[t]
    \centering
    \resizebox{0.8\linewidth}{!}{
    \begin{tabular}{L{4cm}C{2cm}}
        \toprule
        Method & $D^2_{\text{GED}}$ ($\downarrow$) \\
        \midrule
        Temporal Perceiver$^\dagger$~\cite{tan2023temporal} & 45.5 \\
        SC-Transformer$^\dagger$~\cite{li2022structured} & 36.7 \\
        BasicGEBD$^\dagger$~\cite{zheng2024rethinking} & 37.0 \\
        EfficientGEBD$^\dagger$~\cite{zheng2024rethinking} & 38.6 \\
        \rcol \textbf{\ours~(ours)} & \textbf{34.8} \\
        \bottomrule
    \end{tabular}
    }
    \caption{\textbf{Generalized energy distance on Kinetics-GEBD.} $\dagger~$Models with dagger marks are reproduced.}
    \label{tab:ged}
    \vspace{-4mm}
\end{table}
\begin{figure}[t!]
     \centering
     \resizebox{0.8\linewidth}{!}{
     \includegraphics[width=\linewidth]{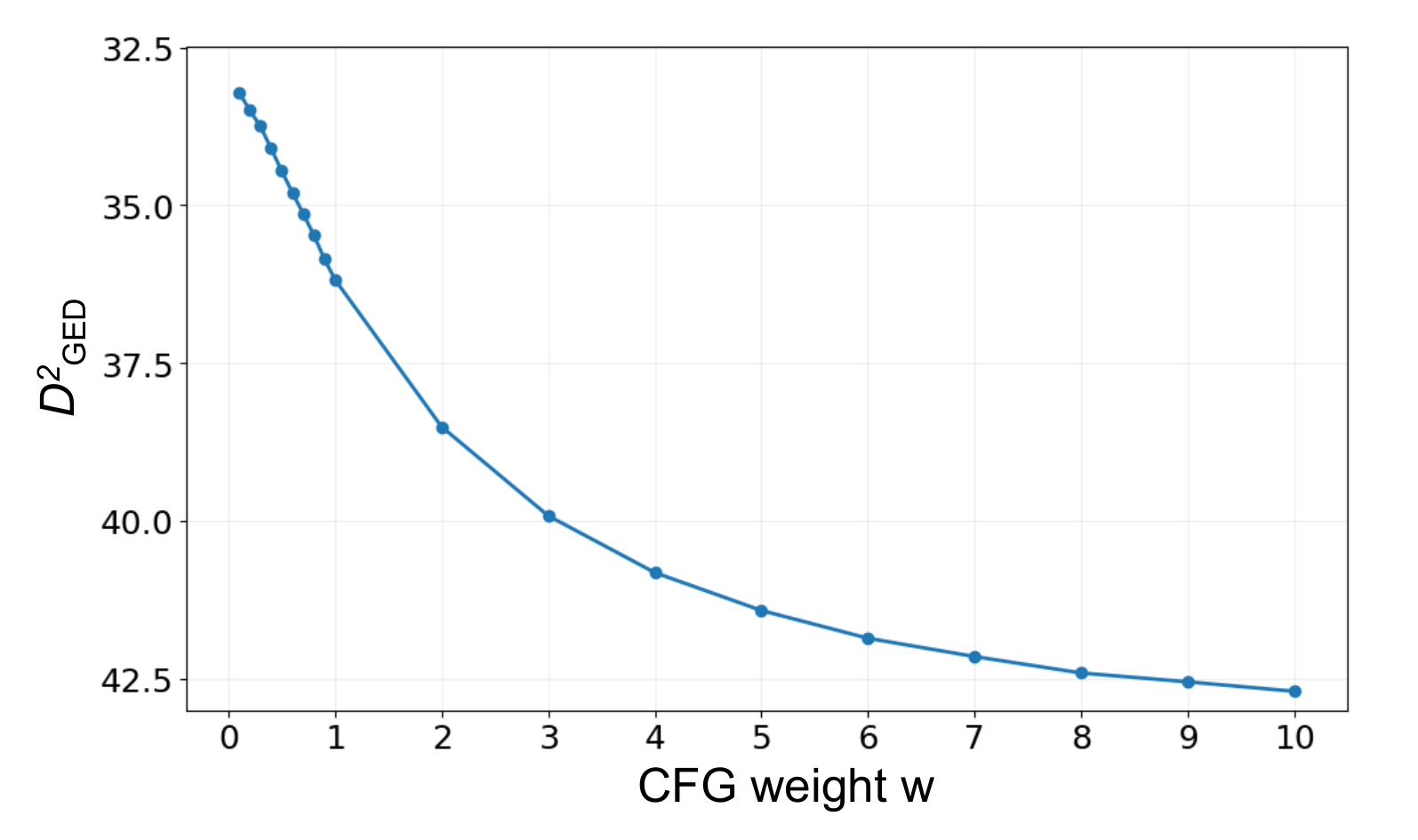}
     }
     \caption{ \textbf{Effects of CFG weight $w$ on GED.} 
     }
     \label{fig:ged_cfg}
     \vspace{-3mm}
\end{figure}
\paragraph{Generalized energy distance (GED).}
\label{sec:ged}
Additionally, we employ the Generalized Energy Distance ($D_{\text{GED}}^2$)~\cite{bellemare2017cramer, salimans2018improving, SZEKELY20131249} on the diversity-aware evaluation. $D_{\text{GED}}^2$ measures the discrepancy between the predicted distributions $\bm{\hat{Y}}$ and the ground truth boundary distributions $\bm{Y}$:
\begin{equation}
    D_{\text{GED}}^2(\bm{\hat{Y}}, \bm{Y}) = 2\E[d(\hat{Y}, Y)] - \E[d(\hat{Y}, \hat{Y'})] -\E[d(Y, Y')],
    \label{eq:ged}
\end{equation}
where $d$ is a distance metric, $\hat{Y}, \hat{Y'}$ are independent samples drawn from the predicted distribution $\bm{\hat{Y}}$, and $Y, Y'$ are independent samples drawn from the ground truth distribution $\bm{Y}$. We adopt $d(i, j) = 1 - \text{F1@}\tau(i, j)$ to evaluate boundary matching score, for arbitrary $i$ and $j$. Here, we set $\tau$ to 0.05. A lower GED score indicates better alignment between predicted and ground truth distributions. The detailed computation is provided in Eq.~\ref{eq:ged_detail}.
\begin{table}[t]
    \centering
    \resizebox{\linewidth}{!}{
    \begin{tabular}{lccccccccc}
    \toprule
    \multirow{2}*[-0.5ex]{Method} & \multicolumn{9}{c}{Threshold $\delta$} \\
    \cline{2-10}
     & 0.1 & 0.2 & 0.3 & 0.4 & 0.5 & 0.6 & 0.7 & 0.8 & 0.9 \\
    \midrule
    DDM-Net~\cite{tang2022progressive}&74.9&75.3&75.7&76.3&76.8&76.0&67.1&43.0&16.7\\
    SC-Transformer$^\dagger$~\cite{li2022structured}&70.9&76.3&77.6&77.3&76.2&74.1&69.6&60.3&37.0\\
    EfficientGEBD~\cite{zheng2024rethinking}& 51.2 & 70.5 & 78.3 & 75.5 & 65.2 & 50.4 & 34.7 & 20.2 & 7.9 \\

    \rcol DiffGEBD(Ours)&\textbf{78.3}&\textbf{78.4}&\textbf{78.4}&\textbf{78.4}&\textbf{78.4}&\textbf{78.4}&\textbf{78.4}&\textbf{78.4}&\textbf{78.4}\\
    \bottomrule
    \end{tabular}
    }
    \caption{\textbf{Robustness on threshold $\delta$.} We report F1@0.05 with different thresholds on Kinetics-GEBD. $\dagger$: reproduced from the official code.}
    \label{tab:ans_ths}
    \end{table}

\begin{table}[t]
    \centering
    \resizebox{\linewidth}{!}{
    \begin{tabular}{L{4cm}C{2cm}C{2cm}}
        \toprule
        Method & Reproduced & Paper \\
        \midrule
        Temporal Perceiver$^\dagger$~\cite{tan2023temporal} & 74.9 & 74.8 \\
        SC-Transformer$^\dagger$~\cite{li2022structured} & 77.4 & 77.7 \\
        BasicGEBD$^\dagger$~\cite{zheng2024rethinking} & 76.9 & 76.8 \\
        EfficientGEBD$^\dagger$~\cite{zheng2024rethinking} & 78.3 & 78.3 \\
        \bottomrule
    \end{tabular}
    }
    \caption{\textbf{F1@0.05 of conventional evaluation protocol on Kinetics-GEBD.} $\dagger~$Models with dagger marks are reproduced using official implementations.}
    \label{tab:detF1}
\end{table}

\begin{equation}
\begin{split}
    D_{GED}^2(\bm{\hat{Y}}, \bm{Y}) = & \frac{2}{N_p N_g} \sum_{i=1}^{N_p} \sum_{j=1}^{N_g} d(\hat{Y}_i, Y_j) \\
    & - \frac{1}{{N_p}^2} \sum_{i=1}^{{N_p}} \sum_{j=1}^{N_p} d(\hat{Y}_i, \hat{Y}_j) \\
    & - \frac{1}{{N_g}^2} \sum_{i=1}^{{N_g}} \sum_{j=1}^{N_g} d(Y_i, Y'_j)
\end{split}
\label{eq:ged_detail}
\end{equation}

Following the diversity-aware evaluation protocol, we evaluate our model using $D_{\text{GED}}^2$. As presented in Table~\ref{tab:ged}, the results demonstrate that our predicted distributions closely align with the ground truth boundary distributions. Figure~\ref{fig:ged_cfg} shows that the CFG weight $w$ increases, the $D_{\text{GED}}^2$ increases, indicating that stronger guidance reduces the diversity of predictions and leads to larger discrepancy between predicted and ground truth distributions.
\begin{table}[t]
    \centering
    \resizebox{0.9\linewidth}{!}{
    \begin{tabular}{ccccc}
        \toprule
        CFG weight $w$ & $\text{F1}_{\text{sym}}$ & $\text{F1}_{\text{p2g}}$ & $\text{F1}_{\text{g2p}}$ & Diversity \\
        \midrule
        0.1 & 73.45 & 74.24 & 73.14 & \textbf{24.64} \\
        0.2 & 73.63 & 74.60 & 73.16 & 23.64 \\
        0.3 & 73.79 & 74.94 & 73.17 & 22.75 \\
        0.4 & 73.87 & 75.19 & 73.11 & 21.88 \\
        0.5 & 73.93 & 75.42 & 73.03 & 21.09 \\
        0.6 & \textbf{73.96} & 75.60 & 72.92 & 20.38 \\
        0.7 & \textbf{73.96} & \textbf{75.76} & 72.79 & 19.73 \\
        0.8 & 73.93 & 75.87 & 72.65 & 19.24 \\
        0.9 & 73.91 & 75.97 & 72.53 & 18.57 \\
        1.0 & 73.85 & 76.04 & 72.37 & 18.07 \\
        2.0 & 73.42 & 76.42 & 71.25 & 14.91 \\
        3.0 & 73.02 & 76.48 & 70.49 & 13.28 \\
        4.0 & 72.73 & \textbf{76.49} & 69.97 & 12.31 \\
        5.0 & 72.52 & 76.48 & 69.58 & 11.70 \\
        6.0 & 72.35 & 76.44 & 69.32 & 11.29 \\
        7.0 & 72.21 & 76.39 & 69.12 & 11.02 \\
        8.0 & 72.08 & 76.34 & 68.93 & 10.82 \\
        9.0 & 72.00 & 76.31 & 68.80 & 10.69 \\
        10.0 & 71.91 & 76.26 & 68.68 & 10.60 \\
        \bottomrule
    \end{tabular}
    }
    \caption{\textbf{Numerical results of the effect of CFG weight $w$ (Fig.~\ref{fig:guidance}).}
    }
    \label{tab:cfg_full}
\end{table}

\paragraph{Robustness on threshold $\delta$.}
In boundary detection, the final boundary prediction $\hat{\bm{y}}_0$ for each frame is thresholded by $\delta$ to determine whether it is classified as a boundary. To assess the robustness of the predictions, we vary $\delta$ from 0.1 to 0.9. As shown in Table~\ref{tab:ans_ths}, \ours~maintains consistently strong performance across different threshold values, demonstrating the robustness of the predicted boundaries.

\paragraph{Reproduced results of previous methods.}
For diversity-aware evaluation protocol, we conduct 5 independent runs for each model. As shown in Table~\ref{tab:detF1}, the average performances of the conventional protocol closely match the reported performance in previous methods, validating the reproducibility.

\begin{table}[t]
    \centering
    \resizebox{0.8\linewidth}{!}{
    \begin{tabular}{ccccc}
        \toprule
        $N_\mathrm{G}$  & $\text{F1}_{\text{sym}}$ & $\text{F1}_{\text{p2g}}$ & $\text{F1}_{\text{g2p}}$ & Diversity \\
        \midrule
        1 & 70.9 & 73.9 & 68.8 & 15.1 \\
        2 & 72.1 & 74.3 & 70.6 & 17.6 \\
        3 & 73.5 & 75.5 & 72.1 & 18.4 \\
        \rcol 4 & \textbf{74.0} & \textbf{75.6} & \textbf{72.9} & 20.4 \\
        5 & 73.0 & 74.4 & 72.4 & \textbf{22.9} \\
        \bottomrule
    \end{tabular}
    }
    \caption{\textbf{Numerical results of the effect of number of annotations (Fig.~\ref{fig:annotator}).}}
    \label{tab:ann_full}
    \vspace{-3mm}
\end{table}

\paragraph{Numerical results of CFG weight $w$.}
Table~\ref{tab:cfg_full} presents the complete numerical results in Fig.~\ref{fig:guidance} in the main paper. While the main paper visualizes these results as plots for better trend analysis, we provide the exact values here for reference. 

\paragraph{Numerical results of number of annotations.}
The numerical results of Fig.~\ref{fig:annotator} are presented in Table~\ref{tab:ann_full}. 

\paragraph{Full results on the diversity-aware evaluation.}
We provide full results with Rel. Dis. threshold ranging from 0.05 to 0.5 on the diversity-aware evaluation protocol. Table~\ref{tab:reldis_diverse} presents the $\text{F1}_\text{sym}$ performance across all thresholds. \ours~outperforms previous methods across all Rel. Dis. thresholds.

\paragraph{Full results on the conventional evaluation.}
We provide full results with Rel. Dis. threshold ranging from 0.05 to 0.5 on the conventional evaluation protocol. 
Table~\ref{tab:kinetics} and Table~\ref{tab:tapos} show the results of Kinetics-GEBD and TAPOS, respectively.

\section{More Example Results}
\label{sec_sup:more_example}
\begin{figure*}[!t]
     \centering
     \resizebox{\linewidth}{!}{
     \includegraphics[width=\textwidth]{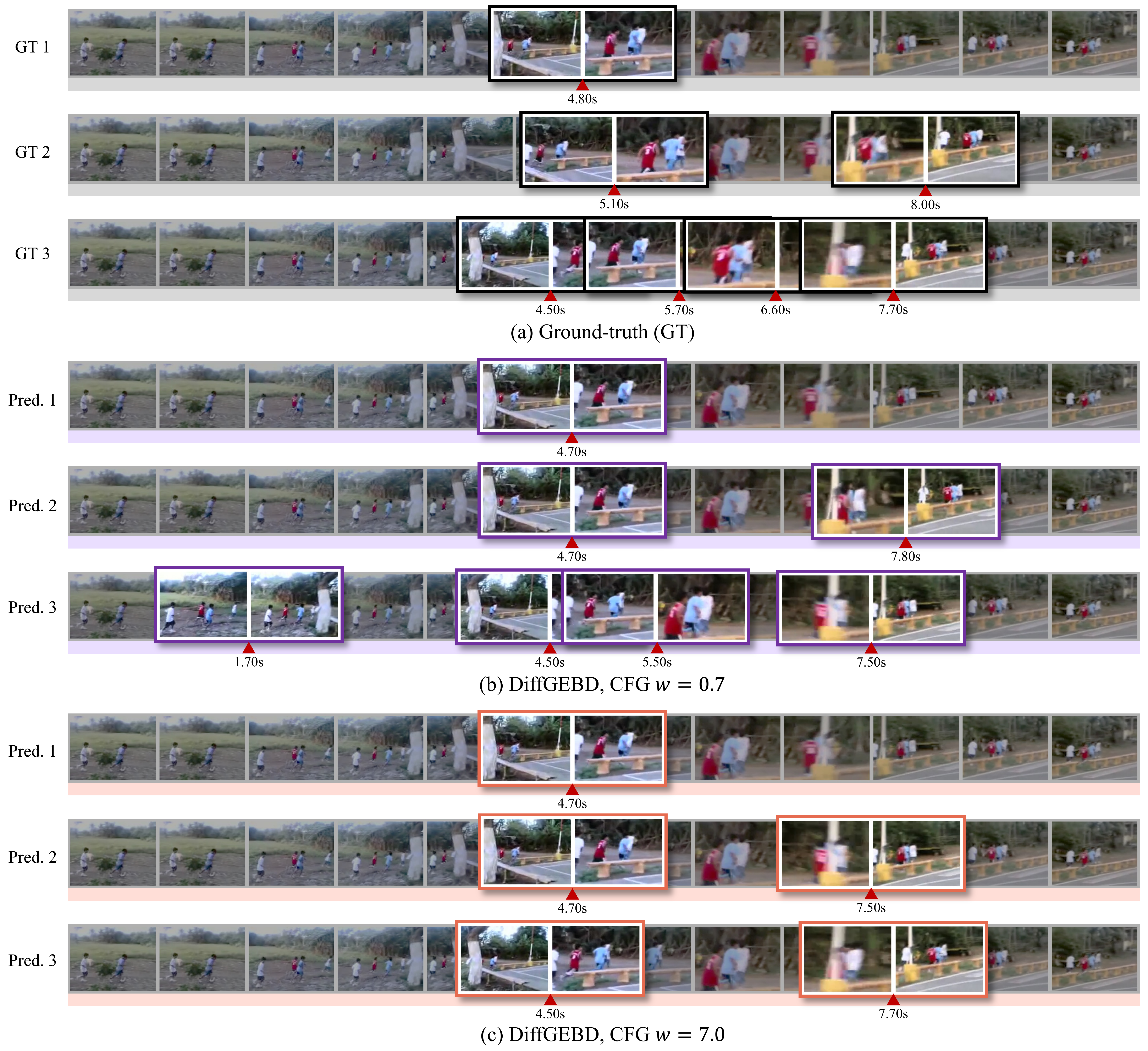}
     } 
     \vspace{-7mm}
     \caption{\textbf{Example results on Kinetics-GEBD.}
     The figure illustrates (a) Ground truth annotations, (b) predictions with $w = 0.3$, and (c) predictions with $w = 7.0$.
     }
     \label{fig_sup:qual}
     \vspace{-6mm}
\end{figure*}

We provide additional qualitative results in Fig~\ref{fig_sup:qual}. The model demonstrates robust detection of boundaries with significant scene changes across all guidance weights. However, for subtle transitions, such as minor object movements observed at 1.70s (\ref{fig_sup:qual}b), the model becomes less sensitive to these boundaries at higher weights (\ref{fig_sup:qual}c). This suggests that lower guidance weights enable the model to capture ambiguous boundaries through its stochastic generation process.

\section{Discussion}
\label{sec_sup:discussion}
\paragraph{Limitations and future work.}
While our diffusion-based method effectively generates multiple predictions, its iterative process significantly slows down inference. Future work will address this limitation by adapting methods like Flow Matching \cite{lipmanflow} and Consistency Models \cite{song2023consistency}. These approaches can achieve high-quality results with a single sampling step, directly addressing the speed limitations.

\paragraph{Broader impact.}
To the best of our knowledge, this work presents the first generative formulation of generic event boundary detection, along with a novel evaluation framework for multiple predictions scenario. We believe that our approach opens up new possibilities for for addressing inherent human ambiguity in event boundaries and provides a new paradigm for future research in this direction.
\begin{table*}[t!]
    \centering
    \scalebox{0.9}{
    \begin{tabular}{l c c c c c c c c c cc}
    \toprule
    \multirow{2}*[-0.5ex]{Method} & \multicolumn{11}{c}{$\text{F1}_{\text{sym}}$ @ Rel. Dis.} \\
    \cmidrule[0.3pt]{2-12}
    &0.05&0.1&0.15&0.2&0.25&0.3&0.35&0.4&0.45&0.5&avg. \\
    \midrule
    Temporal Perceiver$^\dagger$~\cite{tan2023temporal}&69.4&76.9&79.3&80.7&81.6&82.2&82.6&83.0&83.3&83.5&80.2\\
    SC-Transformer$^\dagger$~\cite{li2022structured}& \underline{72.9}&\underline{80.7}&\underline{83.1}&\underline{84.5}&\underline{85.3}&85.9&86.4&86.7&87.0&87.2&\underline{84.0}\\
    BasicGEBD$^\dagger$~\cite{zheng2024rethinking}&72.2&79.7&82.2&83.6&84.6&85.2&85.6&86.0&86.2&86.5&83.2\\
    EfficientGEBD$^\dagger$~\cite{zheng2024rethinking}&72.6&80.3&82.8&84.3&\underline{85.3}&\underline{86.0}&\underline{86.5}&\underline{86.9}&\underline{87.2}&\underline{87.5}&83.9\\
    \rcol DiffGEBD (ours)&\textbf{74.0}&\textbf{81.8}&\textbf{84.2}&\textbf{85.5}&\textbf{86.4}&\textbf{87.0}&\textbf{87.4}&\textbf{87.8}&\textbf{88.1}&\textbf{88.4}&\textbf{85.1}\\
    \bottomrule
    \end{tabular}
    } 
    \caption{
    \textbf{Diversity-aware evaluation on Kinetics-GEBD with Rel.Dis. threshold from 0.05 to 0.5.} We report $\text{F1}_\textbf{sym}$ score varying different relative distance thresholds. Bold numbers indicate the best score, while underlined numbers represent the second-best performance.
    }
    \label{tab:reldis_diverse}
\end{table*}

\begin{table*}[t]
    \centering
    \scalebox{0.9}{
    \begin{tabular}{l c c c c c c c c c cc}
    \toprule
    \multirow{2}*[-0.5ex]{Method} & \multicolumn{11}{c}{F1 @ Rel. Dis.} \\
    \cmidrule[0.3pt]{2-12}
    &0.05&0.1&0.15&0.2&0.25&0.3&0.35&0.4&0.45&0.5&avg. \\
    \midrule
    BMN~\cite{Lin_2019_ICCV}&18.6&20.4&21.3&22.0&22.6&23.0&23.3&23.7&23.9&24.1&22.3\\
    BMN-StartEnd~\cite{Lin_2019_ICCV}&49.1&58.9&62.7&64.8&66.0&66.8&67.4&67.8&68.1&68.3&64.0\\
    TCN~\cite{lea2016segmental}&58.8&65.7&67.9&69.1&69.8&70.3&70.6&70.8&71.0&71.2&68.5\\
    PC~\cite{shou2021generic}&62.5&75.8&80.4&82.9&84.4&85.3&85.9&86.4&86.7&87.0&81.7\\
    SBoCo~\cite{kang2022uboco}&73.2&82.7&85.3&87.7&88.2&89.1&89.4&89.9&89.9&90.7&86.6\\
    Temporal Perceiver~\cite{tan2023temporal}&74.8&82.8&85.2&86.6&87.4&87.9&88.3&88.7&89.0&89.2&86.0\\
    DDM-Net~\cite{tang2022progressive}&76.4&84.3&86.6&88.0&88.7&89.2&89.5&89.8&90.0&90.2&87.3\\
    CVRL~\cite{li2022end}&74.3&83.0&85.7&87.2&88.0&88.6&89.0&89.3&89.6&89.8&86.5\\
    LCVS~\cite{zhang2024local}&76.8&84.8&87.2&88.5&89.2&89.6&89.9&90.1&90.3&90.6&87.7\\
    SC-Transformer~\cite{li2022structured} &{77.7}&{84.9}&{87.3}&{88.6}&{89.5}&{90.0}&{90.4}&{90.7}&{90.9}&{91.1}&{88.1}\\
    BasicGEBD~\cite{zheng2024rethinking}&76.8&83.4&85.7&87.1&87.9&88.5&88.8&89.1&89.4&89.6&86.6\\
    EfficientGEBD~\cite{zheng2024rethinking}&78.3&\underline{85.1}&\underline{87.4}&\underline{88.7}&\underline{89.6}&\underline{90.1}&\underline{90.5}&\underline{90.8}&\underline{91.1}&\underline{91.3}&\underline{88.3}\\
    DyBDet~\cite{zheng2025fine}&\textbf{79.6}&\textbf{85.8}&\textbf{88.0}&\textbf{89.3}&\textbf{90.1}&\textbf{90.7}&\textbf{91.1}&\textbf{91.5}&\textbf{91.7}&\textbf{91.9}&\textbf{89.0}\\

    \rcol DiffGEBD (ours)&\underline{78.4}&84.8&86.8&87.9&88.6&89.1&89.4&89.7&89.9&90.1&87.5\\ 
    \bottomrule
    \end{tabular}
    } 
    \caption{
    \textbf{Comparison with the state of the art on Kinetics-GEBD}. We report F1 score varying different relative distance thresholds. The numbers in boldface indicate the highest score.
    \ours~shows the competitive performance on overall metrics.
    }
    \label{tab:kinetics}
\end{table*}
\begin{table*}[t!]
    \centering
    \scalebox{0.9}{
    \begin{tabular}{l c c c c c c c c c cc}
    \toprule
    \multirow{2}*[-0.5ex]{Method} & \multicolumn{11}{c}{F1 @ Rel. Dis.} \\
    \cmidrule[0.3pt]{2-12}
    &0.05&0.1&0.15&0.2&0.25&0.3&0.35&0.4&0.45&0.5&avg. \\
    \midrule
    ISBA~\cite{ding2018weakly}&10.6&17.0&22.7&26.5&29.8&32.6&34.8&36.9&38.2&39.6&30.2\\
    TCN~\cite{lea2016segmental}&23.7&31.2&33.1&33.9&34.2&34.4&34.7&34.8&34.8&34.8&33.0\\
    CTM~\cite{huang2016connectionist}&24.4&31.2&33.6&35.1&36.1&36.9&37.4&38.1&38.3&38.5&35.0\\
    TransParser~\cite{shao2020intra}&23.9&38.1&43.5&47.5&50.0&51.4&52.7&53.4&54.0&54.5&47.4\\
    PC~\cite{shou2021generic}&52.2&59.5&62.8&64.7&66.0&66.6&67.2&67.6&68.0&68.4&64.3\\
    Temporal Perceiver~\cite{tan2023temporal}&55.2&66.3&71.3&73.8&75.7&76.5&77.4&77.9&\textbf{78.4}&\textbf{78.8}& 73.2\\
    DDM-Net~\cite{tang2022progressive}&60.4&68.1&71.5&73.5&74.7&75.3&75.7&76.0&76.3&76.7&72.8\\
    SC-Transformer~\cite{li2022structured}&61.8&69.4&72.8&74.9&76.1&76.7&77.1&77.4&77.7&78.0&74.2\\
    BasicGEBD~\cite{zheng2024rethinking}&60.0&66.6&-&-&-&73.1&-&-&-&74.8&71.0\\
    EfficientGEBD~\cite{zheng2024rethinking}&63.1&70.5&-&-&-&\textbf{77.4}&-&-&-&78.6&74.8\\
    DyBDet~\cite{zheng2025fine}&62.5&70.1&73.4&75.6&\textbf{76.7}&77.2&\textbf{77.5}&\textbf{77.9}&78.1&78.4&74.7\\
     \rcol DiffGEBD (ours)&\textbf{65.8}&\textbf{71.8}& \textbf{74.1}&\textbf{75.7}&\underline{76.4}&77.0&\underline{77.4}& 77.7&78.0&78.1&\textbf{75.2}\\
    \bottomrule
    \end{tabular}
    } 
    \caption{
    \textbf{Comparison with the state of the art on TAPOS.} We report F1 score varying different relative distance thresholds. The numbers in boldface indicate the highest score. \ours~shows the state-of-the-art performance on overall metrics.}
    \label{tab:tapos}
\end{table*}
\clearpage

\end{document}